\documentclass[11pt]{article}
\usepackage[margin=1in]{geometry}
\usepackage{times}
\usepackage{natbib}

\usepackage{amsmath,amsfonts,bm}

\def\eqref#1{(\ref{#1})}

\def\1{\bm{1}}

\DeclareMathAlphabet{\mathsfit}{\encodingdefault}{\sfdefault}{m}{sl}
\SetMathAlphabet{\mathsfit}{bold}{\encodingdefault}{\sfdefault}{bx}{n}

\usepackage{graphicx}        
\usepackage{booktabs}        
\usepackage[hypertexnames=false,hidelinks,pdfborder={0 0 0}]{hyperref}
\usepackage{url}
\usepackage{xcolor}          
\usepackage{pifont}          
\usepackage{amssymb}         
\usepackage{fontawesome5}    
\usepackage{algorithm}
\usepackage{algpseudocode}
\usepackage{tabularx}
\usepackage{wrapfig}
\usepackage{float}            
\usepackage{capt-of}
\usepackage{subfiles}
\usepackage{etoc}\usepackage{xspace}            

\brokenpenalty=10000 \newcommand{\liberoindustry}{\textsc{LIBERO}-Industry\xspace}

\newcommand{\cmark}{\textcolor{green!70!black}{\ding{51}}} 
\newcommand{\xmark}{\textcolor{red!80!black}{\ding{55}}}

\newcommand{\rwTODO}[1]{\textcolor{red!80!black}{[TODO: #1]}}
\algrenewcommand\algorithmicrequire{\textbf{Input:}}
\algrenewcommand\algorithmicensure{\textbf{Output:}}

\title{LIBERO-Industry: Benchmarking Robotic Manipulation under Industrial Process Constraints}

\author{%
Ming Shin Cindy Sung$^{1}$, Duan Leijiahao$^{1}$,
Haoxiang Wang$^{1}$, Boning Tan$^{1}$,\\
Liangna Fu$^{1}$, Chao Tan$^{1}$, Yafu Li$^{2}$,
Yulong Chen$^{3,4}$,\\
Chaoqun Wan$^{1}$, Zhen Huang$^{1}$,
Shuxian Liang$^{1*}$, Xian-Sheng Hua$^{1}$\\[0.5em]
\normalsize $^{1}$Institute of AI for Engineering, Tongji University \quad
$^{2}$The Chinese University of Hong Kong\\$^{3}$University of Aberdeen \quad $^{4}$University of Cambridge\\[0.3em]
\href{https://IAIE-PhysicalAI.github.io/LIBERO-Industry}{\textcolor{blue}{https://IAIE-PhysicalAI.github.io/LIBERO-Industry}}
}
\date{}

\begin{document}

\etocdepthtag.toc{main}

\maketitle
\begingroup
\renewcommand{\thefootnote}{\fnsymbol{footnote}}
\footnotetext[1]{Corresponding author}
\endgroup

\begin{abstract}
    \noindent Industrial manipulation in flexible manufacturing, e.g., high-mix, low-volume (HMLV) electronics assembly, requires robot policies to adapt to changing products and production conditions.
    Such settings impose industrial process constraints, including product variants, moving material flows, finite execution windows, and tight geometric tolerances.
    However, existing evaluation frameworks do not systematically capture these constraints when assessing generalist robot policies.
    To address this gap, we introduce \liberoindustry, a novel benchmark built on the widely used robot learning benchmark \textsc{LIBERO}, that translates the aforementioned industrial constraints into controllable task conditions and measurable success criteria.
    Specifically, \liberoindustry comprises three classic task scenarios: Static-to-Dynamic (\textsc{S2D}) loading, Dynamic-to-Static (\textsc{D2S}) unloading, and Dynamic-to-Dynamic (\textsc{D2D}) transfer.
    The suites contain 216 tasks in total that systematically vary object geometry, conveyor speed, interaction-window duration, and geometric acceptance region size.
    Extensive experiments across five generalist robot policies (including vision-language-action models and world-action models) show that performance progressively degrades as industrial constraints are introduced and combined.
    Under a shared fine-tuning and evaluation protocol, the \liberoindustry co-training condition achieved higher observed target-transfer completion than the original-\textsc{LIBERO} condition.
\end{abstract}

\section{Introduction}

Flexible manufacturing, e.g., high-mix, low-volume (HMLV) electronics assembly,  requires frequent adaptation to changing products and production conditions, 
requiring substantial re-engineering effort on conventional automation \citep{gan2023review,schlenoff2021agile}.
Generalist robot policies, including Vision-Language-Action (VLA) models such as RT-2 \citep{brohan2023rt2} and Octo \citep{octo_2023}, offer a learned alternative by conditioning robot actions on multimodal observations and language instructions across diverse tasks.
However, industrial readiness requires more than task-level adaptability, particularly under the precision and operating conditions encountered in industrial settings \citep{li2025transferring}.

\begin{figure}[!t]
    \centering
    \includegraphics[width=0.90\textwidth]{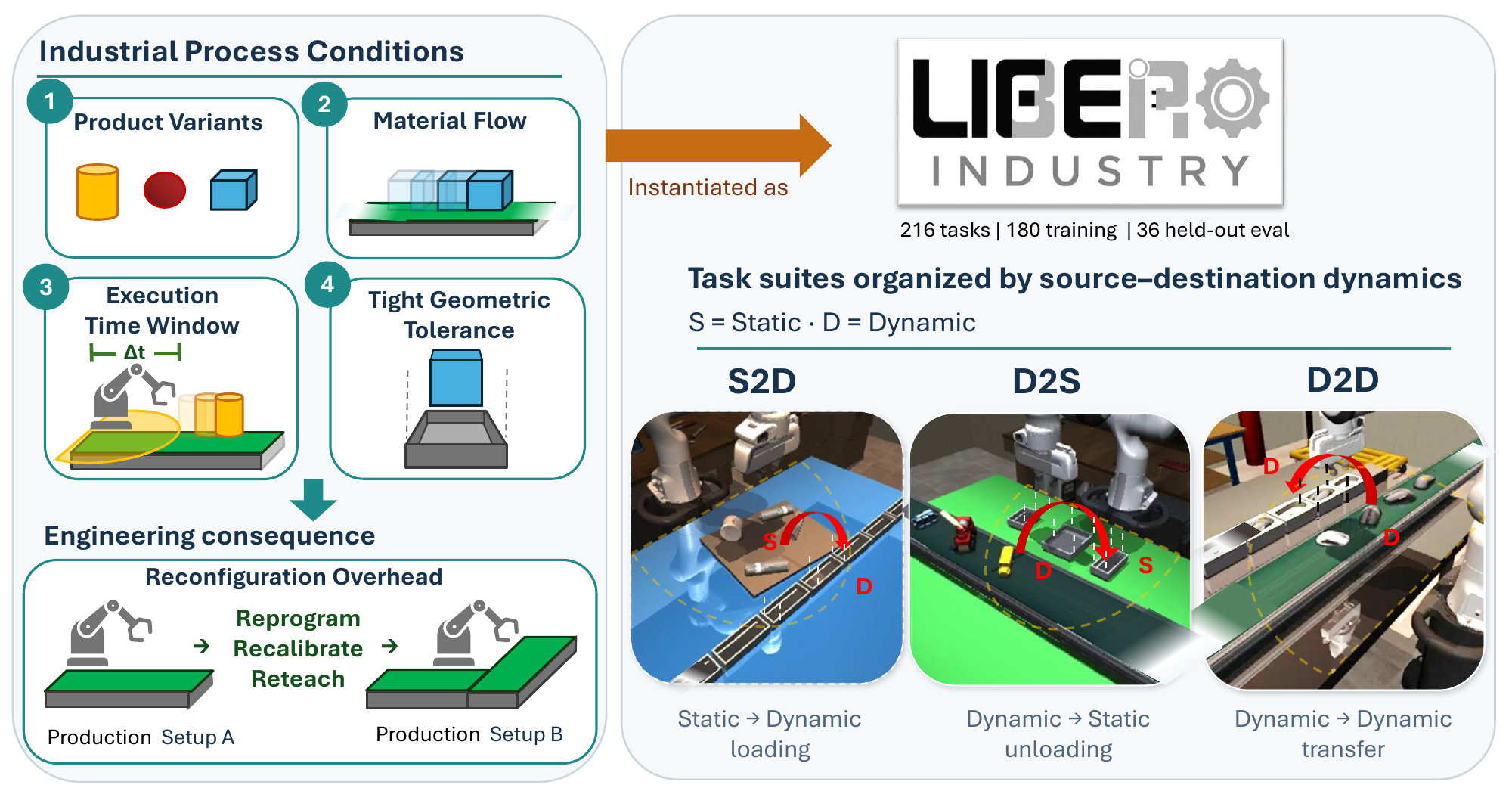}
    \caption{\textbf{\liberoindustry overview.} Motivated by industrial process constraints,
    the benchmark evaluates generalist robot policies across three classic industrial scenarios.}
    \label{fig:teaser}
\end{figure}

Industrial manipulation in flexible production is shaped by industrial constraints, introducing product variants, moving material flows, finite execution windows, and tight geometric tolerances, as illustrated in Figure~\ref{fig:teaser}.
Product variants may require different setups and processing times \citep{gan2023review}, making frequent production changes harder to accommodate with fixed systems \citep{koren1999reconfigurable}.
Moving material flows impose finite time windows for manipulation \citep{islam2021constanttime}, while parts must meet tight geometric tolerances despite manufacturing and perception errors~\citep{jiang2022roboticassembly}.
Traditional automation works reliably when tasks and environments remain stable, but frequent changes often require reprogramming, recalibration, robot re-teaching, or changes to the workcell \citep{schlenoff2021agile,gan2023review} (detailed in Appendix~\ref{app:control_paradigms}, Table~\ref{tab:control_paradigms}). Together, these characteristics define the industrial process constraints that embodied manipulation evaluations should capture.

However, existing embodied-manipulation and manufacturing-oriented benchmarks do not systematically evaluate generalist robot policy performance under industrial constraints, as summarized in Table~\ref{tab:benchmark_comparison}.
On one hand, general-purpose manipulation benchmarks, such as LIBERO, CALVIN, and Meta-World, provide standardized protocols for studying task learning, generalization, and language-conditioned control \citep{liu2023libero,mees2022calvin,Yuetal2019}, yet largely abstract away the timing, motion, and placement requirements imposed by production processes.
On the other hand, manufacturing-oriented research and benchmarks, including the Agile Robotics for Industrial Automation Competition (ARIAC), address robotic flexibility, dynamic production conditions, and operational performance \citep{harrison2018ariac,jiang2022roboticassembly} but do not comprehensively evaluate these industrial constraints or generalist robot policies.
Consequently, how moving sources and destinations, finite execution windows, and tight geometric tolerances affect task success, throughput, and stage-wise failures remains underexplored, particularly as these requirements accumulate across multi-object workloads.

To address this research gap, we introduce \liberoindustry, an extension of
the widely used robot learning benchmark \textsc{LIBERO} \citep{liu2023libero} for evaluating generalist robot policies under industrial process constraints, that operationalizes these constraints as controllable task conditions and measurable success criteria. As shown in Figure~\ref{fig:teaser}, \liberoindustry comprises three task suites derived from typical industrial scenarios: Static-to-Dynamic (\textsc{S2D}) loading, Dynamic-to-Static (\textsc{D2S}) unloading, and Dynamic-to-Dynamic (\textsc{D2D}) transfer. These suites differ in whether workpiece sources and destinations are stationary or moving, capturing different forms of material-flow interaction.
Each suite includes three progressive difficulty levels with controlled variations in workspace geometry, visual conditions, and task configuration, together with structured perturbations informed by industrial operating conditions and standards.
Tasks couple spatial completion requirements with finite execution windows, allowing process constraints to be introduced systematically and in combination.

We evaluate five generalist robot policies, including VLA models and a world-action model (WAM) in simulation, measuring task success, throughput, and stage-wise completion.
We further assess real-world performance through physical D2S trials on an AUBO-i5H arm, comparing fine-tuning effectiveness with \liberoindustry and \textsc{LIBERO}  under matched conditions.

In summary, our main contributions are as follows:
\begin{enumerate}
    \item[(a)] \textbf{Industrial Process Benchmark:} We introduce \liberoindustry, a novel benchmark that systematically evaluates product variants, moving material flows, finite execution windows, and tight geometric tolerances.

    \item[(b)] \textbf{Process-Constrained Task Design:} We translate industrial process requirements as controllable task conditions, including conveyor velocity, bounded execution duration, and geometric completion requirement sizes, enabling their systematic evaluation individually and in combination.

    \item[(c)] \textbf{Empirical Evaluation and Failure Diagnosis:} We benchmark five generalist policies, including VLA models and a WAM, showing progressive performance degradation as industrial constraints are introduced and combined. Under a shared fine-tuning and evaluation protocol, the \liberoindustry co-training condition achieved higher observed target-transfer completion than the original-\textsc{LIBERO} condition.
\end{enumerate}

\section{Related Work}
\label{sec:related}
\begin{table}[t]
    
    \begin{center}
\renewcommand{\arraystretch}{1.45}

\caption{
Comparison of recent embodied manipulation and manufacturing-oriented benchmarks across policy and industrial-process evaluation dimensions.
\cmark\ denotes explicit evaluation, $\sim$ partial or indirect evaluation, and \xmark\ no evaluation. An execution time window denotes an operational constraint on task completion rather than a reported runtime metric.
}
\label{tab:benchmark_comparison}

\setlength{\tabcolsep}{3pt}
\begin{tabularx}{\linewidth}{@{}
>{\raggedright\arraybackslash}p{0.20\linewidth}
>{\raggedright\arraybackslash}p{0.12\linewidth}
*{6}{>{\centering\arraybackslash}X}
@{}}
\toprule

{\footnotesize\textbf{Dataset}}
& {\footnotesize\textbf{Domain}}
& {\footnotesize\textbf{Language-conditioned policy}}
& {\footnotesize\textbf{Product variants}}
& {\footnotesize\textbf{Material flow}}
& {\footnotesize\textbf{Execution time window}}
& {\footnotesize\textbf{Geom. tolerance}}
& {\footnotesize\textbf{Real-world validation}} \\
\midrule

\multicolumn{8}{l}{\textit{General-purpose embodied-AI benchmarks}} \\

LIBERO
& Household
& \cmark
& $\sim$
& \xmark
& \xmark
& \xmark
& \xmark \\

LIBERO-PLUS
& Household
& \cmark
& $\sim$
& \xmark
& \xmark
& \xmark
& \xmark \\

VLA-Arena
& Household
& \cmark
& $\sim$
& \xmark
& \xmark
& \xmark
& \xmark \\

LIBERO-X
& Household
& \cmark
& $\sim$
& \xmark
& \xmark
& \xmark
& \xmark \\

DynamicVLA
& Generalized
& \cmark
& $\sim$
& \xmark
& \xmark
& \xmark
& \cmark \\

DOMINO
& Generalized
& \cmark
& $\sim$
& \xmark
& \xmark
& \xmark
& \cmark \\

\midrule

\multicolumn{8}{l}{\textit{Manufacturing-oriented simulation benchmarks}} \\

ARIAC 2025
& EV battery
& \xmark
& \cmark
& \cmark
& $\sim$
& \xmark
& \xmark \\

AI-CPS
& Industry
& \xmark
& $\sim$
& \xmark
& \xmark
& \xmark
& \xmark \\

RoboBPP
& Logistics
& \xmark
& \cmark
& \cmark
& \xmark
& $\sim$
& $\sim$ \\

\midrule

LIBERO-Industry\newline\textbf{(ours)}
& Industry
& \cmark
& \cmark
& \cmark
& \cmark
& \cmark
& \cmark \\

\bottomrule
\end{tabularx}

    \end{center}
\end{table}
\subsection{Generalist Robot Policies}
\label{subsec:embodied_policies}

Generalist robot policies map multimodal observations and language instructions
to actions across diverse tasks. RT-2 and OpenVLA autoregressively decode
discrete action tokens~\citep{brohan2023rt2,kim2024openvla}, whereas $\pi_0$
and $\pi_{0.5}$ use flow-matching action heads to generate continuous action
chunks~\citep{black2024pi,physicalintelligence2025pi05}. Recent World Action
Models (WAMs) couple future-state prediction with action generation. WorldVLA,
DreamZero, and Cosmos Policy jointly predict future visual states and
actions~\citep{cen2025worldvla,dreamzero2026,kim2026cosmospolicy}. VPP instead
conditions action prediction on predictive visual representations; Fast-WAM
uses video prediction for co-training but directly decodes actions at inference,
while FLARE aligns policy representations with latent embeddings of future
observations~\citep{hu2024vpp,yuan2026fastwam,zheng2025flare}. However, their
behavior under calibrated industrial operating conditions remains underexplored.

\subsection{Generalist Robot Manipulation Benchmarks}
\label{subsec:embodied_benchmarks}

A variety of embodied manipulation benchmarks have been proposed to study multi-task learning, long-horizon control, and knowledge transfer. Meta-World targets multi-task and meta-reinforcement learning, CALVIN evaluates language-conditioned long-horizon execution, and LIBERO studies lifelong knowledge transfer~\citep{Yuetal2019,mees2022calvin,liu2023libero}. This line of work has expanded to robustness under object, scene, viewpoint, physical, and linguistic variations through LIBERO-PRO, LIBERO-PLUS, VLA-Arena, and LIBERO-X~\citep{Zhouetal2025,Feietal2025,Zhangetal2025,Wangetal2026}. More recent benchmarks, including DOMINO and the Dynamic Object Manipulation (DOM) benchmark introduced with DynamicVLA, consider dynamic disturbances and moving-object manipulation~\citep{Fangetal2026,Xieetal2026}. Yet these benchmarks do not formulate production-line operating conditions as controllable evaluation variables for generalist policies. \liberoindustry addresses this gap through conveyor-driven material flows, finite execution windows, task-specific placement predicates, and repeated multi-object handling, together with process-informed perturbations derived from industrial operating conditions.

\subsection{Industrial Process-Aware Evaluation}
\label{subsec:industrial_evaluation}

A variety of manufacturing-oriented benchmarks have been proposed to evaluate production workflows, industrial manipulation, and robotic logistics. ARIAC assesses robotic agility under changing simulated production orders, AI-CPS evaluates industrial manipulation controllers in Isaac Sim, and RoboBPP studies physics-aware online bin packing using industrial data~\citep{nist2025ariac,zhou2024aicps,wang2025robobpp}. Related manufacturing research further characterizes reconfigurability, material transport, cycle-time requirements, and pose accuracy~\citep{gan2023review,groover2014automation,monch2021variabletakt,iso9283_1998}. Nevertheless, neither embodied manipulation benchmarks nor manufacturing-oriented evaluations systematically assess generalist policies under controlled production-line operating conditions. \liberoindustry targets this shared limitation by exposing domain-specific process conditions as controlled benchmark variables across standardized conveyor-transfer tasks.

\section{LIBERO-Industry Benchmark Design}
\label{sec:task_design}

The benchmark aims to evaluate how reliably and efficiently generalist robot policies perform material-transfer tasks in changing industrial workcells.
To support comparable evaluation, each task instance is assigned explicit completion predicates and a predeclared episode horizon derived from its transport schedule.

\subsection{Preliminaries and Task Interface}

We evaluate policies with different architectures and objectives through the language-conditioned observation--action interface used by LIBERO and CALVIN~\citep{liu2023libero,mees2022calvin}:
\begin{equation}
o_t\in\mathcal{O},\qquad
\mathbf{u}_t\sim\pi_\theta(\cdot\mid o_{\leq t},\ell_m),\qquad
\mathbf{u}_t\in\mathcal{U},
\label{eq:policy_interface}
\end{equation}
where $\ell_m$ is the instruction for evaluation episode $m$, $o_t$ contains the benchmark-defined visual and proprioceptive observations, and $\mathbf{u}_t$ is a robot action produced at control frequency $f_c$.
The full simulator state $s_t$ is available to the evaluator for computing task predicates but need not be available to the policy.

We denote the task instance for evaluation episode $m$ by $\mathcal{T}_m$, with $\mathcal{C}_m$ collecting its temporal opportunities and goal specification:
\begin{equation}
\begin{aligned}
\mathcal{T}_m
&=\left\langle \ell_m,\mu_{0,m},\mathcal{W}_m,
\xi_m^{\mathrm{src}},\xi_m^{\mathrm{dst}},\mathcal{C}_m\right\rangle,\\
\mathcal{C}_m
&=\left\langle
\left\{(a_{mj},H_{mj})\right\}_{j=1}^{N_{\text{target},m}},
\mathcal{G}_m
\right\rangle .
\end{aligned}
\label{eq:task_instance}
\end{equation}
Here, $\mu_{0,m}$ is the initial-state specification; $\mathcal{W}_m$ contains the target workpieces and any pre-existing destination workpieces; and $\xi_m^{\mathrm{src}}$ and $\xi_m^{\mathrm{dst}}$ denote the \emph{source process} and \emph{destination process}, respectively, each of which may be static, continuously moving, or intermittent.
For each of the $N_{\mathrm{target},m}$ target workpieces, indexed by $j$, $a_{mj}$ is its availability time and $H_{mj}$ is the end of its physical opportunity horizon.
The set $\mathcal{G}_m$ contains the goal propositions that define valid completion for the target workpieces and, when present, pre-existing destination workpieces.

\subsection{Industrial Process Constraints}

Each task has two process-level conditions:
transport determines how the source and destination evolve, while
temporal opportunity determines when a transfer remains feasible.

\textbf{Transport condition.}
Under a continuous source process, target workpieces move at signed velocity $V_s$; under an intermittent destination process, the destination moves at signed velocity $V_d$ between stationary service periods.
These transport states change the workpiece or goal geometry over time. Speed is therefore an operating variable rather than a success metric.
The continuous and intermittent modes follow established material-transport terminology~\citep{groover2014automation}.

\textbf{Temporal opportunity condition.}
For the continuous source process, target workpieces separated by center-to-center pitch $d_{\text{src}}$ have transport-event interval
\begin{equation}
\tau_{\text{src}} = \frac{d_{\text{src}}}{|V_s|}.
\label{eq:arrival_interval}
\end{equation}
For the pulse-indexed destination process, slot pitch $s_{\text{slot}}$ and stationary service dwell $T_{\text{dwell}}$ define the transport-event interval
\begin{equation}
\tau_{\text{dst}}=T_{\text{dwell}}+
\frac{s_{\text{slot}}}{|V_d|}.
\label{eq:destination_interval}
\end{equation}
These intervals define the physical transport schedule, while $[a_{mj},H_{mj}]$ bounds the opportunity during which workpiece $j$ can be completed. For the selected source arrival intervals, see Appendix~\ref{app:timing_protocol}.

\begin{figure*}[!t]
    \centering
    \includegraphics[width=\textwidth]{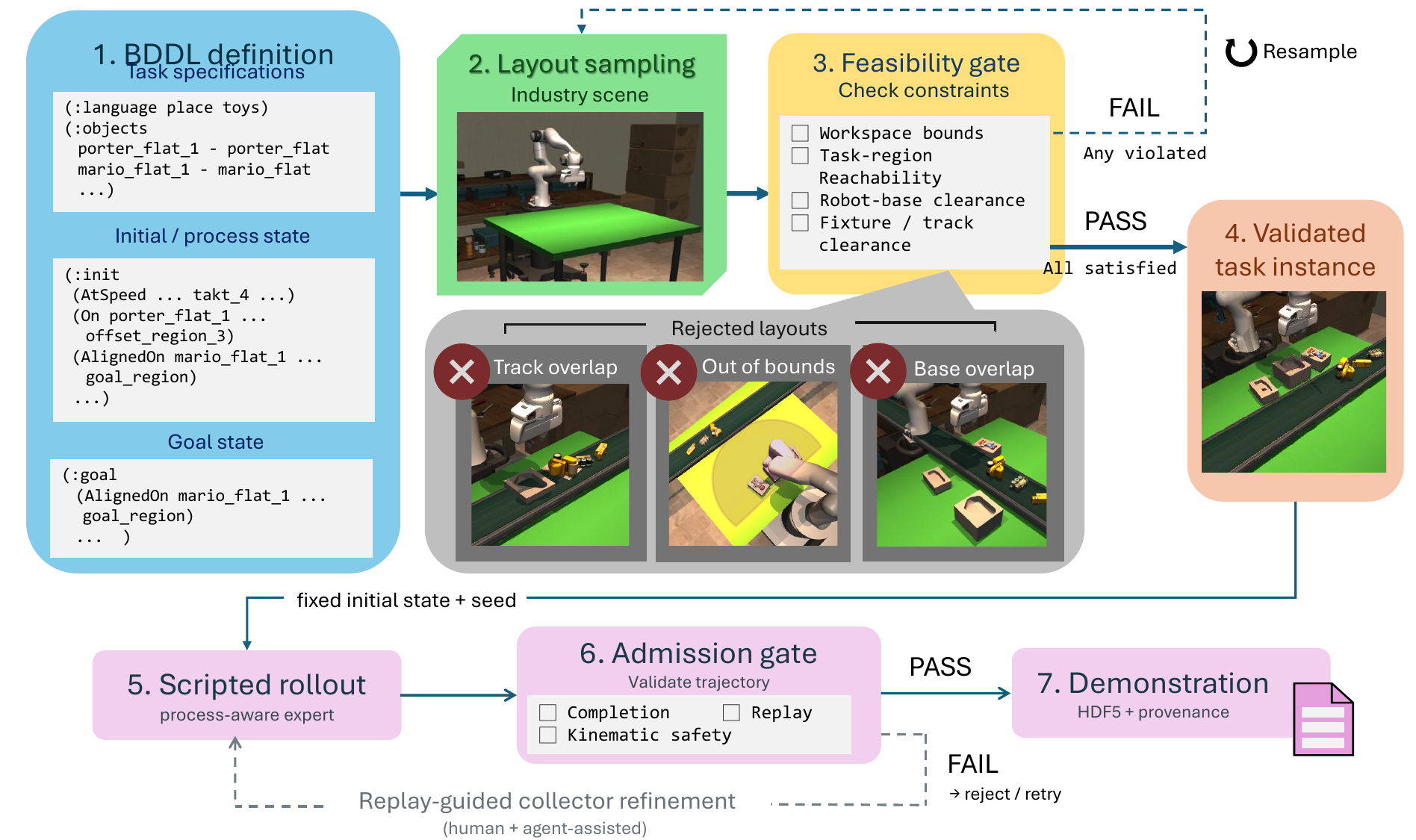}
    \caption{
    \textbf{Task generation and scripted demonstration collection in \liberoindustry.}
    BDDL specifies objects, process states, and goals; sampled layouts pass a feasibility gate or are resampled (top).
    A task-specific scripted expert attempts a rollout from a fixed initial state and seed (bottom).
    The diagram shows completion, kinematic-safety, and replay checks before accepted trajectories are stored as HDF5 demonstrations with provenance.
    Failed attempts can prompt replay-guided collector refinement.
    }
    \label{fig:task-generation-pipeline}
\end{figure*}

\subsection{Process-Aware Procedural Task Generation}
\label{subsec:procedural_generation}

Figure~\ref{fig:task-generation-pipeline} illustrates the task generation pipeline.
While workspace support follows LIBERO's region-based formulation~\citep{liu2023libero}, our generator additionally conditions object initialization and task timing on material-flow dynamics.

Given an operation anchor, static fixtures are sampled with level-specific pose perturbations; moving sources are back-projected using transport direction, velocity, arrival interval, and offset; moving destinations set slot spacing from geometry and travel direction and service intervals from velocity and dwell (Appendix~\ref{app:catalog_distribution}).
Each layout is serialized as BDDL with its initial predicates, transport conditions, instruction, and goals, and infeasible layouts are resampled before instantiation.
\subsection{Task Suites and Difficulty Levels}
\label{subsec:generalization_tiers}

The source and destination transport states produce three non-static transfer suites, summarized in Table~\ref{tab:task_taxonomy}.

\begin{table}[t]
\centering
\setlength{\tabcolsep}{3pt}
\renewcommand{\arraystretch}{0.9}
\caption{
Task taxonomy and dataset composition of \liberoindustry across the three workflow suites.
Full combinations are defined over object, conveyor speed, motion direction, and difficulty level.
For D2S, four objects are selected as a stratified subset from six available objects.
}
\label{tab:task_taxonomy}
\begin{tabular}{@{}llcccc@{}}
\toprule
\textbf{Suite} & \textbf{Workflow logic}
& \multicolumn{1}{c}{\textbf{Full combos}}
& \textbf{Instantiated} & \textbf{Train} & \textbf{Test} \\
\cmidrule(lr){3-3}
& & \textbf{Obj. $\times$ Speed $\times$ Dir. $\times$ Lvl.} & & & \\
\midrule
D2S & Conveyor Belt $\rightarrow$ Precision Slot
& 4 $\times$ 3 $\times$ 2 $\times$ 3 & 72 & 60 & 12 \\
S2D & Material Plate $\rightarrow$ Dynamic Slot
& 4 $\times$ 3 $\times$ 2 $\times$ 3 & 72 & 60 & 12 \\
D2D & Conveyor Belt $\rightarrow$ Dynamic Slot
& 4 $\times$ 3 $\times$ 2 $\times$ 3 & 72 & 60 & 12 \\
\midrule
\textbf{Total} & & & 216 & 180 & 36 \\
\bottomrule
\end{tabular}
\end{table}

\begin{itemize}
    \item \textbf{Dynamic-to-Static (D2S) unloading}: moving source $\rightarrow$ stationary fixture.
    The policy must acquire a continuously transported workpiece before its source opportunity ends and then satisfy the fixture-specific completion predicate.

    \item \textbf{Static-to-Dynamic (S2D) loading}: stationary source $\rightarrow$ intermittently moving destination.
    The policy must place a workpiece during an available destination service period and maintain the required placement as the carrier resumes motion.

    \item \textbf{Dynamic-to-Dynamic (D2D) transfer}: moving source $\rightarrow$ intermittently moving destination.
    The policy must satisfy both the source-acquisition opportunity and the destination-service opportunity before stable placement can be credited.
\end{itemize}

The benchmark instantiates these suites with workpieces from multiple industrial domains and level-specific destination fixtures (for asset inventory see Appendix~\ref{app:asset_taxonomy}).

For task instance $m$, the benchmark generator converts the instantiated transport schedule and completion allowance into a rollout budget:
\begin{equation}
T_{\text{roll},m}=t_{\text{start},m}+T_{\text{event},m}+T_{\text{complete},m}+t_{b,m},
\qquad
K_{\text{roll},m}=\left\lceil f_cT_{\text{roll},m}\right\rceil ,
\label{eq:episode_horizon}
\end{equation}
where $t_{\text{start},m}$ is the initialization or first-arrival offset, $T_{\text{event},m}$ is the time spanned by the relevant transport events, $T_{\text{complete},m}$ is the final manipulation or service allowance, and $t_{b,m}$ is the terminal buffer.
Appendix~\ref{app:timing_protocol} expands this expression for D2S, S2D, and D2D and reports the fixed allowances for every suite, level, and transport condition.
These values are shared by all evaluated policies and are benchmark time limits rather than universal factory takt times.

Each level preserves the same high-level transfer objective while jointly increasing completion precision and environmental variation: Level~1 pairs a broad tray region with the nominal workcell condition, Level~2 adds object-matched grooves with prescribed orientations and bounded perturbations, and Level~3 combines form-fitting inlays with broader perturbations.
For moving destinations, success additionally requires carrier retention during transport, with perturbations specified using the pose-accuracy and pose-repeatability terminology of ISO~9283~\citep{iso9283_1998} (see Secs.~\ref{app:kinematic_predicates}, \ref{app:kinematic_physics}, and~\ref{app:environmental_specs}).

\subsection{Evaluation Protocol and Metrics}
\label{sec:evaluation_protocol}

We report three metrics: binary task success rate and two
partial completion rates.

\paragraph{Primary metric: success rate (SR).}
For each $g\in\mathcal{G}_m$, let $E_m(g)\in\{0,1\}$ denote its Boolean
evaluation at the predeclared rollout boundary $K_{\text{roll},m}$.
Episode success and Task Success Rate over $M$ evaluation
episodes are
\begin{equation}
Y_m=\prod_{g\in\mathcal{G}_m}E_m(g),
\qquad
\operatorname{SR}=\frac{1}{M}\sum_{m=1}^{M}Y_m .
\label{eq:task_success_rate}
\end{equation}

\paragraph{Complementary metrics: partial completion rate (PCR).}
Partial Completion Rates measure progress toward the transfer objective.
For episode $m$, let $N_{\text{transferred},m}$ denote the number of target
workpieces credited by the transfer predicate, $N_{\text{base},m}$ the number
of pre-existing destination workpieces, and $N_{\text{slotted},m}$ the number
of target and pre-existing workpieces satisfying their required terminal
predicates. The episode-level scores are
\begin{equation}
\mathrm{PCR}_{\mathrm{T},m}
=
\frac{N_{\text{transferred},m}}{N_{\text{target},m}},
\qquad
\mathrm{PCR}_{\mathrm{S},m}
=
\frac{N_{\text{slotted},m}}
{N_{\text{target},m}+N_{\text{base},m}} .
\label{eq:partial_completion}
\end{equation}
The first score isolates progress on target transfers, whereas the second also
penalizes displacement of pre-existing destination workpieces. We report their
episode means:
\begin{equation}
\mathrm{PCR}_{\mathrm{T}}
=
\frac{1}{M}\sum_{m=1}^{M}\mathrm{PCR}_{\mathrm{T},m},
\qquad
\mathrm{PCR}_{\mathrm{S}}
=
\frac{1}{M}\sum_{m=1}^{M}\mathrm{PCR}_{\mathrm{S},m}.
\label{eq:mean_partial_completion}
\end{equation}

\section{Experiments}
\label{sec:experiments}

Following the experimental setup (Section~\ref{subsec:experimental_setup}), we evaluate overall benchmark performance (Section~\ref{subsec:results}), analyze how performance changes as industrial stressors accumulate and where failures concentrate (Section~\ref{subsec:ablations}), and assess whether co-training with \liberoindustry simulation data improves physical D2S performance and robustness over co-training with the original \textsc{LIBERO} data (Section~\ref{sec:real_validation}).

\subsection{Experimental Setup}
\label{subsec:experimental_setup}

We follow the original \textsc{LIBERO} simulation settings except that each rollout is bounded by the task-specific horizon $K_{\mathrm{roll},m}$ (Eq.~\eqref{eq:episode_horizon}). We evaluate five publicly available \textsc{LIBERO}-finetuned checkpoints: $\pi_{0.5}$, OpenVLA-OFT, UniVLA, X-VLA, and Cosmos Policy. The deterministic evaluation set contains 36 tasks---12 each from S2D, D2S, and D2D---with four tasks per difficulty level in each suite and 10 initial states per task.
We report SR, $\mathrm{PCR}_{\mathrm{T}}$, and $\mathrm{PCR}_{\mathrm{S}}$ following Section~\ref{sec:evaluation_protocol} (details in Appendix~\ref{app:model_details}).

\begin{table}[!t]
\centering
\caption{\textbf{Evaluation-set results.} Scores in \%; 360 rollouts per policy. $\mathrm{PCR}_{\mathrm{S}}$ includes initially valid destination workpieces that remain valid.}
\label{tab:main_results}
\begin{tabular}{@{}lccc@{}}
\toprule
Policy & SR & $\mathrm{PCR}_{\mathrm{T}}$ & $\mathrm{PCR}_{\mathrm{S}}$ \\
\midrule
$\pi_{0.5}$ & 0.0 & 0.4 & 12.5 \\
OpenVLA-OFT & 0.0 & 0.0 & 11.2 \\
UniVLA & 0.0 & 0.0 & 12.2 \\
X-VLA & 0.0 & 0.0 & 11.9 \\
Cosmos Policy & 0.0 & 0.0 & 11.8 \\
\bottomrule
\end{tabular}
\end{table}

\subsection{Benchmark Results}
\label{subsec:results}
Table~\ref{tab:main_results} reports performance across the 36 held-out tasks. All five checkpoints were evaluated over 360 rollouts. None satisfied the complete-task success predicate, yielding an SR of 0.0\% (0/360) for every policy. The suite-level breakdown likewise shows zero SR in S2D, D2S, and D2D (details in Appendix~\ref{app:evaluation_set_breakdown}).
Among the evaluated policies, $\pi_{0.5}$ exhibits the strongest observed partial-completion profile. It is the only policy with nonzero target-transfer completion ($\mathrm{PCR}_{\mathrm{T}}=0.4\%$ overall, including 1.1\% in S2D)  and also records the highest terminal-slot occupancy ($\mathrm{PCR}_{\mathrm{S}}=12.5\%$, compared with 11.2--12.2\% for the remaining policies). The held-out BDDL specifications yield a mean initial-state slotting baseline of
$\mathrm{PCR}_{\mathrm{S}}^{\mathrm{init}}=12.31\%$
($12.3\%$ at the reporting precision of Table~\ref{tab:main_results};
see Appendix~\ref{app:evaluation_set_breakdown}).
Thus, $\mathrm{PCR}_{\mathrm{S}}$ values near this baseline largely reflect
pre-existing destination workpieces that remain valid, whereas newly completed
target transfers are isolated by $\mathrm{PCR}_{\mathrm{T}}$.

\begin{figure}[!t]
    \centering
    \includegraphics[width=0.90\columnwidth]{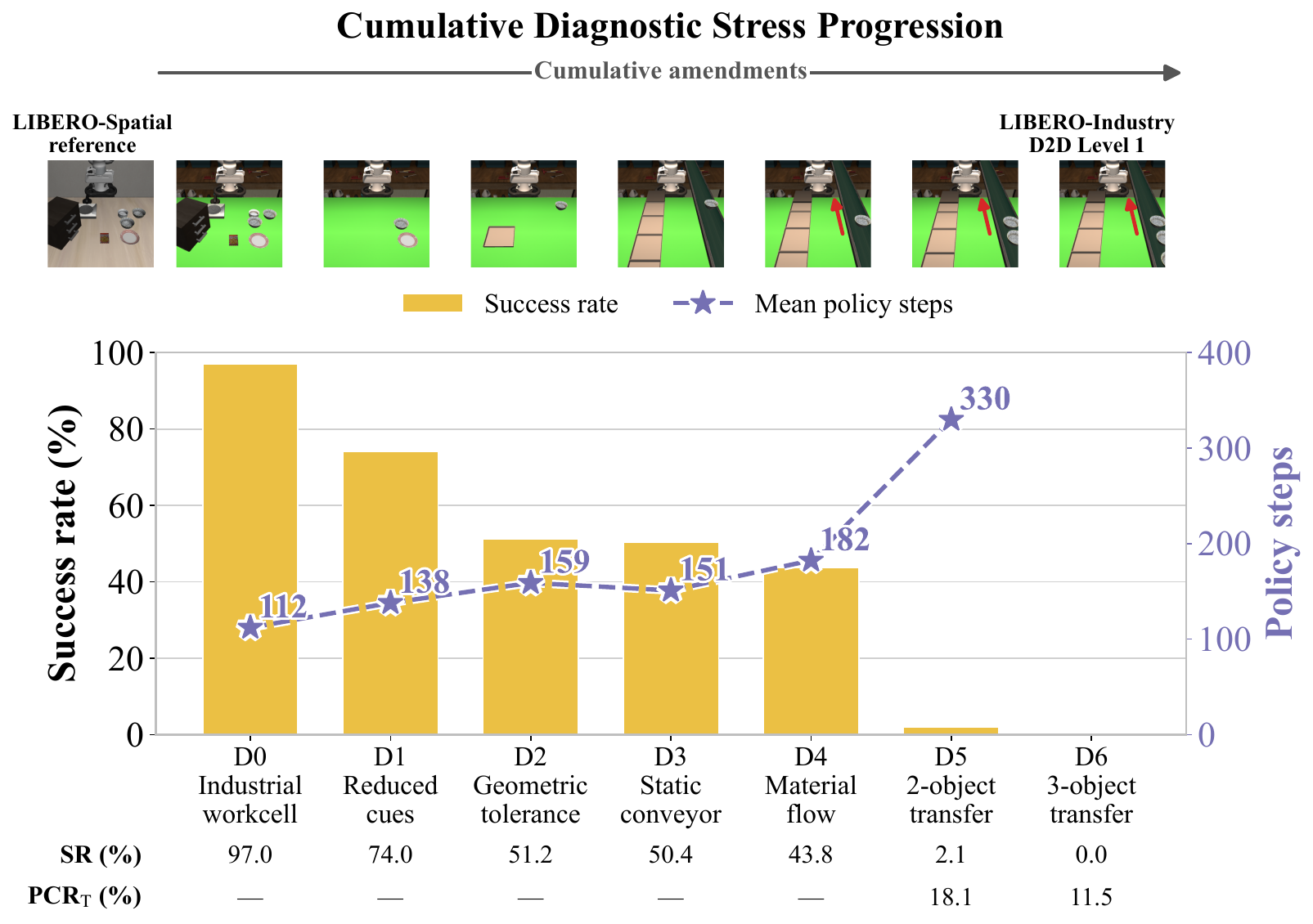}
    \caption{
    \textbf{Cumulative diagnostic stress progression for $\pi_{0.5}$.}
    Each bar denotes complete-task success of the stress settings for n trials (D0--D1: $n=200$; D2--D6: $n=240$), and $\mathrm{PCR}_{\mathrm{T}}$ denotes mean target-object transfer completion.  Stars show mean policy steps over successful trajectories; D6 has no successful trajectories.
    The progression is diagnostic rather than causal.
    }
    \label{fig:ablation-progression}
\end{figure}

\subsection{Ablation and Findings}
\label{subsec:ablations}
Figure~\ref{fig:ablation-progression} traces the cumulative D0--D6 stress progression for $\pi_{0.5}$. SR declines from 97.0\% at D0 to 43.8\% at D4, then collapses to 2.1\% at D5 and 0.0\% at D6 as the task expands from one to multiple objects. Because each stage retains the preceding modifications, the progression is diagnostic rather than causal. Partial transfer persists at D5--D6 ($\mathrm{PCR}_{\mathrm{T}}=18.1\%$ and $11.5\%$), while reach-stage failures account for 132/240 and 202/240 episodes, respectively, localizing the principal bottleneck to longer multi-object sequences (Appendix~\ref{app:diagnostic_stress_progression}; Appendix~\ref{app:representative_failure_trajectories}).

\begin{figure}[!t]
    \centering
    \includegraphics[width=\textwidth]{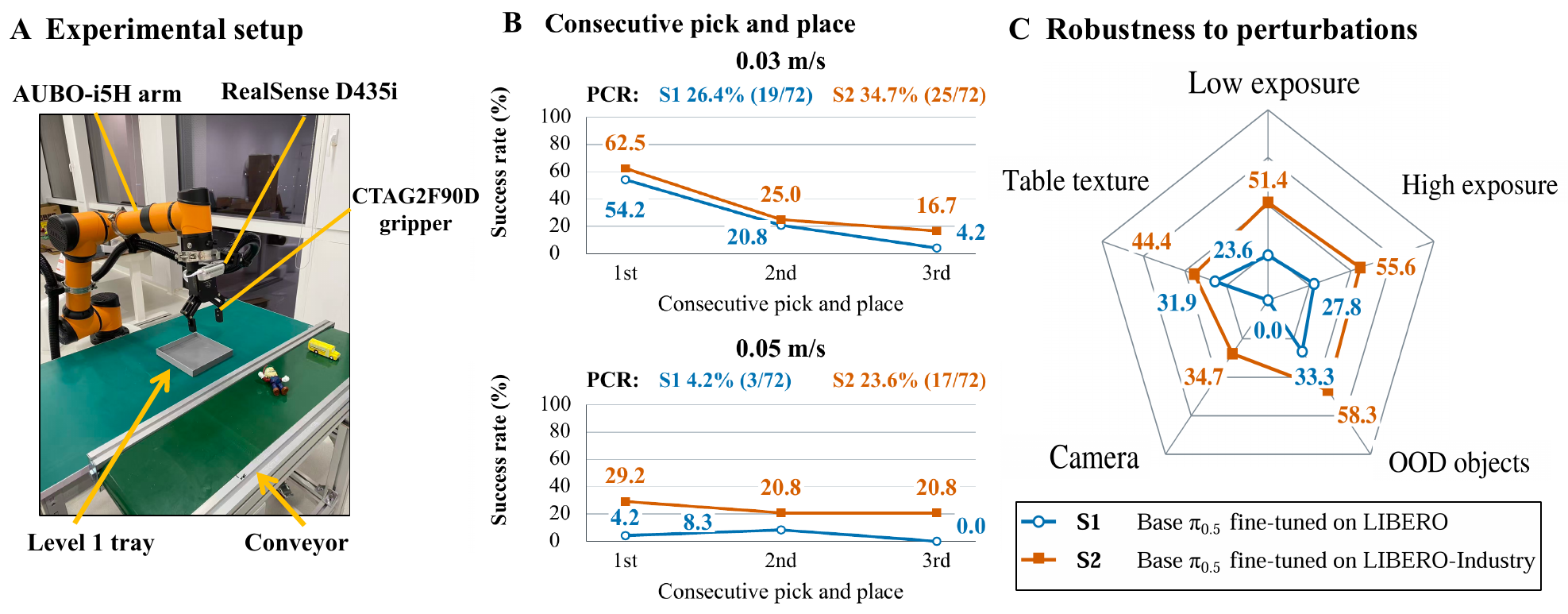}
    \caption{
    \textbf{Real-world validation.}
    \textbf{A:} AUBOi5H--CTAG2F90D D2S workcell.
    \textbf{B:} Per-step release-on-tray success for \textbf{S1} and \textbf{S2} at conveyor speeds of 0.03 and 0.05\,m/s over 24 trials per condition; the table reports aggregate target-transfer completion as $\mathrm{PCR}_{\mathrm{T}}$ out of 72 planned placements.
    \textbf{C:} Target-transfer completion ($\mathrm{PCR}_{\mathrm{T}}$) under five physical operating-condition variations, with exact values labeled.
    }
    \label{fig:real-world-validation-summary}
\end{figure}

\subsection{Real-World Experiments}
\label{sec:real_validation}

\providecommand{\rwTODO}[1]{\textbf{[TODO: #1]}}

We evaluate held-out physical D2S unloading on an AUBOi5H--CTAG2F90D workcell. Each trial requires the policy to transfer three workpieces sequentially from a moving conveyor into a stationary Level~1 tray. The evaluation uses four three-workpiece combinations and two held-out slot permutations per combination, yielding eight held-out object-order configurations. At each conveyor speed, each policy is evaluated over 24 trials, corresponding to 72 planned placements per policy--speed condition. The held-out combinations and permutations are disjoint from the physical training split (Appendix~\ref{app:real_world_dataset}; Table~\ref{tab:real_world_permutation_split}).

We compare two sim--real co-training setups initialized from the same $\pi_{0.5}$ base checkpoint. \textbf{S1} combines 1,724 original-\textsc{LIBERO} simulation demonstrations with 95 physical D2S demonstrations, whereas \textbf{S2} combines 1,412 \liberoindustry simulation demonstrations with the same 95 physical demonstrations. Both use the same 9:1 frame-based simulation-to-real sampling ratio, 30,000 training steps, model configuration, and physical evaluation protocol. The comparison therefore isolates the simulation-data source and corpus size while holding the physical supervision and training procedure fixed (Appendix~\ref{app:rw_training}; Table~\ref{tab:rw_training_config}).

Figure~\ref{fig:real-world-validation-summary}B reports target-transfer completion at conveyor speeds of 0.03 and 0.05\,m/s. At 0.03\,m/s, \textbf{S2} achieves $\mathrm{PCR}_{\mathrm{T}}=34.7\%$ (25/72), compared with 26.4\% (19/72) for \textbf{S1}; at 0.05\,m/s, the corresponding values are 23.6\% (17/72) and 4.2\% (3/72). The fraction of trials completing the third placement is also higher for \textbf{S2}: 16.7\% versus 4.2\% at 0.03\,m/s and 20.8\% versus 0.0\% at 0.05\,m/s. Across both aggregate transfer completion and later placement steps, \textbf{S2} maintains higher observed performance, with the largest separation at the faster conveyor speed.

Figure~\ref{fig:real-world-validation-summary}C evaluates changes in illumination, object appearance, camera configuration, and table texture. \textbf{S2} exceeds \textbf{S1} across all five evaluated variations by 12.5--34.7 percentage points, with the largest difference under the camera variation (34.7\% versus 0.0\%). This consistent ordering indicates that the observed advantage of \textbf{S2} extends beyond the nominal D2S configuration under the tested physical operating-condition changes.
\section{Conclusion and future work}
\label{sec:conclusion}

We introduced \liberoindustry, a benchmark for evaluating generalist policies under industrial process constraints, including product variability, material flows, finite execution-time windows, and tight geometric tolerance.
Built on \textsc{LIBERO}, the benchmark organizes these conditions into S2D , D2S and D2D task suites, with controlled variations in task configuration and operating conditions. Our results show performance degradation across the cumulative industrial stress progression and higher D2S manipulation ability and generalization for the \liberoindustry co-training condition than for the original-\textsc{LIBERO} condition.
A key future direction is to determine whether generalist policies can retain flexibility without sacrificing industrial production rates; the benchmark's finite opportunity windows abstract this requirement from heterogeneous factory needs. Future evaluation should move toward production-calibrated settings that relate cycle times to empirical operating rates and test repeated material arrivals, recovery events, and long-horizon throughput. Ultimately, industrial readiness should be assessed not only by task completion but by compatibility with diverse production processes.

\bibliography{references}
\bibliographystyle{plainnat}


\clearpage
\appendix

\etocdepthtag.toc{appendix}

\begingroup

\color{black}
\setlength{\parindent}{0pt}
\setlength{\parskip}{0pt}

\etocsettagdepth{main}{none}
\etocsettagdepth{appendix}{subsection}
\etocsetnexttocdepth{subsection}

\hypersetup{
    hidelinks,
    linktoc=all
}

\thispagestyle{empty}

\vspace*{0.2cm} 

{
    \centering

    {\LARGE\bfseries
    LIBERO-Industry: Benchmarking Robotic Manipulation
    under Industrial Process Constraints
    \par}

    \vspace{0.35cm}  

    {\Large Appendix\par}
}

\vspace{0.55cm} 
\etocsettocstyle
    {%
        {\large\bfseries Contents\par}
        \vspace{0.2cm}  
    }
    {}

\normalsize
\linespread{1.0}\selectfont  
\tableofcontents

\endgroup
\clearpage
\raggedbottom 



\section{Control Paradigms and Technical Comparison}
\label{app:control_paradigms}

This section provides a structured comparison of traditional, learning-based, and generalist robotic control paradigms. Table~\ref{tab:control_paradigms} outlines the trade-offs in model scale, data dependencies, and physical precision limits that motivate the transition toward high-fidelity simulation environments like \liberoindustry.

\begin{table}[H]
\centering
\caption{Comparative Analysis of Control Paradigms in Industrial Manipulation Tasks.}
\label{tab:control_paradigms}
\small
\renewcommand{\arraystretch}{1.10}
\setlength{\tabcolsep}{4pt}
\begin{tabularx}{\textwidth}{
  >{\raggedright\arraybackslash}p{2.3cm} 
  >{\raggedright\arraybackslash}p{2.8cm} 
  >{\raggedright\arraybackslash}p{2.0cm} 
  >{\raggedright\arraybackslash}p{2.2cm} 
  >{\raggedright\arraybackslash}X
}
\toprule
\textbf{Paradigm} & \textbf{Representative Arch. / Algos} & \textbf{Model Size \& Freq.} & \textbf{Data / Training Source} & \textbf{Typical Precision \& Repeatability$^\ddagger$} \\ \midrule

\textbf{Traditional Rule-Driven} & 
FSM / BT; A* / TAMP; PID, Force, Hybrid Position--Force, Impedance &
No learned params; 1000--2000~Hz & 
Zero data; physical models \& expert kinematics & 
Typically sub-millimeter repeatability in calibrated industrial setups \\ \hline

\textbf{Modular Perception-Action} & 
YOLOv8 / FastSAM + LLM-TAMP (MoveIt / OMPL) & 
Perception: 100M--1B+ ($<10$~Hz); Planner: Det. ($>200$~Hz) & 
Zero downstream trajectories; pre-trained vision \& 3D CAD models & 
Reported millimeter-scale alignment, depending on perception and calibration \\ \hline

\textbf{Imitation Learning (IL)} & 
ACT, Diffusion Policy, DP3 & 
50M--100M params; 30--100~Hz inference & 
Behavioral cloning from 50--100 expert demonstrations & 
Sub-millimeter performance reported for structured task-specific settings \\ \hline

\textbf{Model-Free / Residual RL} & 
PPO, SAC, OEC-IRRL, RL-100 & 
$<20$M params; $>500$~Hz & 
Online/Offline RL via environment reward signals & 
Task- and reward-dependent; sub-millimeter alignment reported in selected settings \\ \hline

\textbf{Generalist Robot Policies} & 
VLA models: RT-2, OpenVLA; generalist diffusion/transformer policies: Octo, RDT-1B & 
1B--7B+ params; 3--10~Hz & 
Multi-task or multi-embodiment pre-training with language, image-goal, or action conditioning & 
Centimeter-scale errors reported in selected precision-placement evaluations \\ \hline

\textbf{World Action Models (WAM)} & 
DreamZero, 1XWM, UniSim & 
14B+ params; 3--7~Hz closed-loop control & 
Joint video-action diffusion / latent generative dynamics & 
Task-dependent; no standardized physical-repeatability measure \\ \bottomrule
\end{tabularx}
\par\smallskip
\parbox{\textwidth}{%
  \footnotesize
  $^\ddagger$ Precision and repeatability summarize representative reported results and depend on the robot, sensing and calibration pipeline, task geometry, and evaluation protocol.
}
\end{table}

As detailed in Table~\ref{tab:control_paradigms}, the landscape of robotic manipulation control can be broadly categorized into distinct evolutionary tiers, each exhibiting specific deployment bottlenecks when transitioned to High-Mix Low-Volume (HMLV) manufacturing:

\begin{enumerate}    \item \textbf{Classical and Modular Control Stacks.}    Classical robot systems separate task sequencing, motion planning,    trajectory generation, and feedback control. Finite-state machines    and behavior trees encode process logic, while geometric or    task-and-motion planners generate feasible motions. PID, force,    hybrid position--force, and impedance controllers then regulate    trajectory tracking and physical interaction    \citep{garrett2021integrated,whitney1977force,    raibert1981hybrid,hogan1985impedance}.    These systems can provide high-rate feedback and repeatable    execution under calibrated conditions, but their performance    depends on accurate models, engineered task logic, and careful    integration of perception, planning, and control.    \item \textbf{Task-Specific Learning Policies.}    Imitation Learning methods such as ACT and Diffusion Policy learn    observation-to-action mappings from expert demonstrations, whereas    model-free Reinforcement Learning optimizes behavior through    task-specific reward signals    \citep{zhao2023act,chi2023diffusion,haarnoja2018soft}.    These approaches reduce the need to manually specify every motion    or contact strategy and can represent temporally coordinated or    multimodal behaviors. However, their performance remains closely    tied to the training task, embodiment, observation--action    interface, reward design, and coverage of demonstrations or    environment interactions.    \item \textbf{Generalist Robot Policies and World Action Models.}    Generalist robot policies use multi-task or multi-embodiment    pretraining to support language-conditioned behavior and transfer    across tasks. VLAs such as RT-2 and OpenVLA directly connect visual    and language inputs to robot actions, while Octo and RDT-1B use    generalist transformer or diffusion-policy formulations    \citep{brohan2023rt2,kim2024openvla}.    World Action Models additionally predict future visual dynamics    together with action sequences, providing an explicit generative    representation of how the scene may evolve    \citep{dreamzero2026}. Their broader semantic and task coverage is    promising, although runtime frequency, contact behavior, and    physical precision continue to depend on the model architecture,    inference hardware, action representation, and downstream    adaptation.\end{enumerate}


\section{Benchmark Design and Task Specification}
\label{app:benchmark_specification}

\subsection{Task Suites, Assets, and Fixtures}
\label{app:asset_taxonomy}

Each instantiated task is encoded as a BDDL specification containing its objects, initial conditions, language instruction, and goal propositions.
The propositions in $\mathcal{G}_m$ are grounded by the corresponding benchmark predicates.

\begin{figure}[H]
    \centering
    \includegraphics[width=\textwidth]{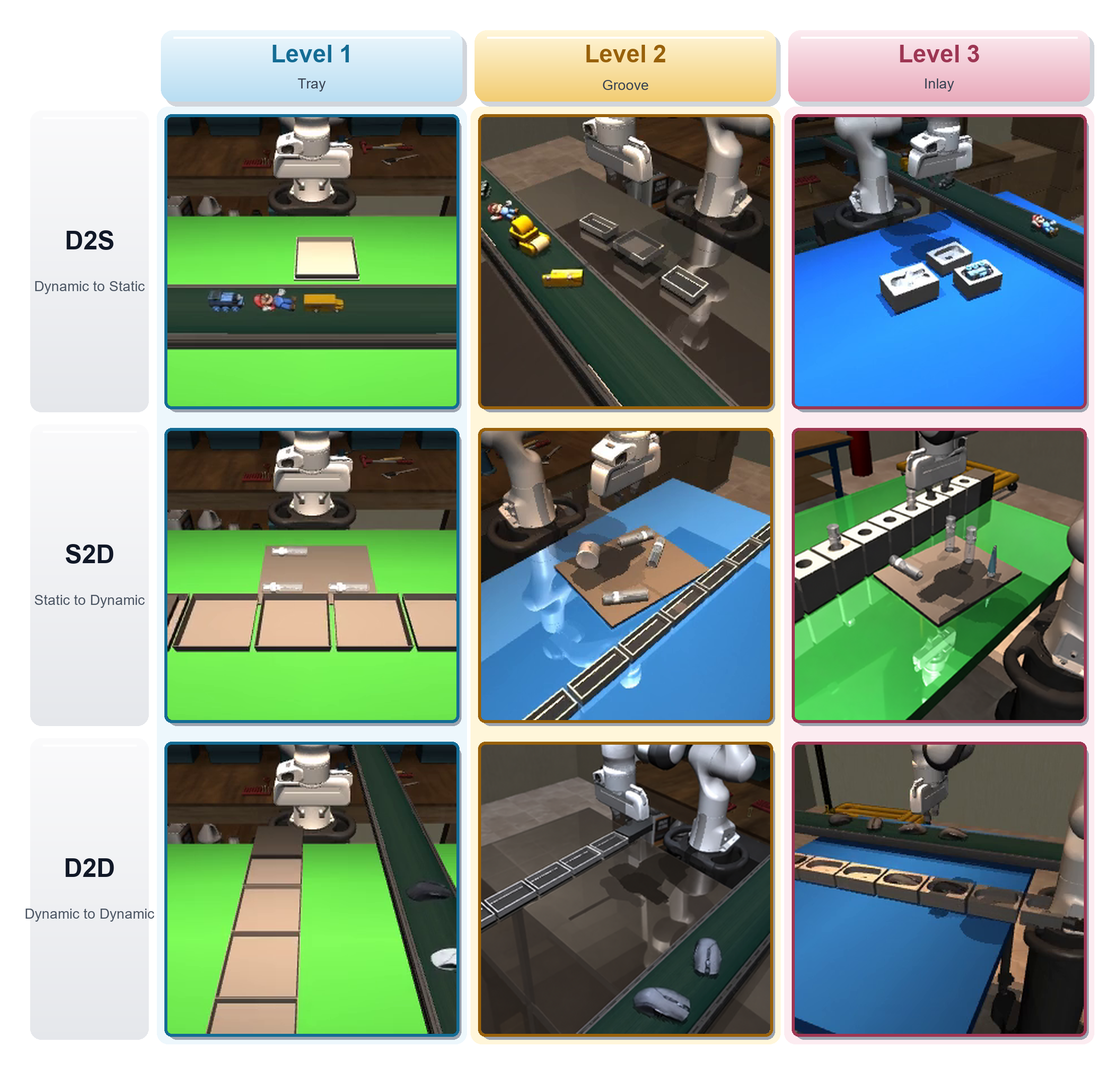}
    \caption{Representative \liberoindustry task instantiations.
    Rows show Dynamic-to-Static (D2S), Static-to-Dynamic (S2D), and
    Dynamic-to-Dynamic (D2D) transport. Columns progress from Level~1 sector
    trays to Level~2 object-specific grooves and Level~3 form-fitting inlays.
    Header and border colors identify difficulty levels and do not encode
    measured performance.}
    \label{fig:task_instantiation_grid}
\end{figure}

Figure~\ref{fig:task_instantiation_grid} visualizes the suite--level matrix
with one representative task from each combination.

We introduce a collection of industrial object assets spanning several manufacturing domains. The base models are adapted from publicly accessible and commercial asset repositories, including an industrial conveyor model~\citep{ConveyorAssemblyCGTrader}, Google Scanned Objects~\citep{Downs2022GoogleScannedObjects}, and commercial design configurations~\citep{CosmeticsPackCGTrader}. Table~\ref{tab:appendix_assets} catalogs the validated workpiece inventory and the corresponding destination fixtures.

\begin{figure}[H]
    \centering
    \includegraphics[width=\textwidth]{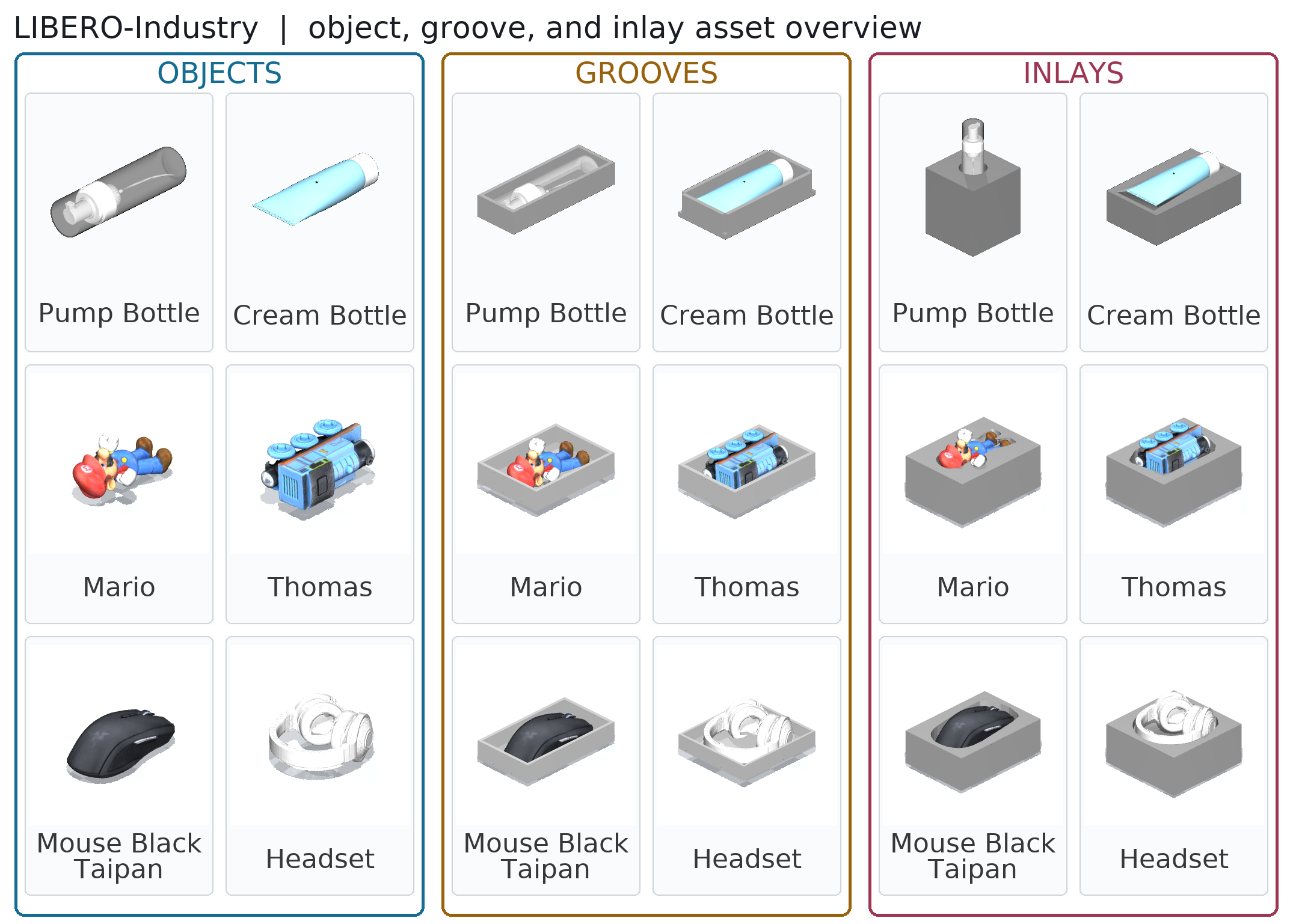}
    \caption{Representative \liberoindustry workpiece and fixture
    assets. From left to right, the panels show six evaluation workpieces and
    their object-specific Level~2 grooves and Level~3 form-fitting inlays,
    spanning personal care, model figurines and vehicles, and consumer-electronics
    domains.}
    \label{fig:appendix_asset_triptych}
\end{figure}

We enforce a strict \textbf{1~cm geometric clearance} rule across three difficulty tiers:
\begin{itemize}
    \item \textbf{Sector Trays (Level 1):} Domain-wide general receptacles dimensioned to fit the maximum bounding box of the largest workpiece within that industrial category plus a 1~cm margin. 
    \item \textbf{Grooves (Level 2):} Object-specific slots generated strictly from the bounding box dimensions of individual workpieces plus a 1~cm margin.
    \item \textbf{Form-Fitting Inlays (Level 3):} Object-specific negative-space profiles modeled via proportional 1cm boolean mesh expansion in Blender.
\end{itemize}

We process inlay configurations through the \texttt{obj2mjcf} utility using Volumetric Hierarchical Approximate Convex Decomposition (V-HACD) to decompose hollow geometries into convex sub-meshes \cite{obj2mjcf}.

\begin{table}[H]
\caption{Systematic Breakdown of Novel Evaluation Workpieces and Target Mating Fixtures Ordered by Insertion Difficulty}
\label{tab:appendix_assets}
\begin{center}
\footnotesize
\setlength{\tabcolsep}{3pt}
\begin{tabularx}{\linewidth}{@{}>{\raggedright\arraybackslash}p{0.19\linewidth}>{\raggedright\arraybackslash}X>{\raggedright\arraybackslash}X@{}}
\hline
\textbf{Industrial Domain} & \textbf{Base Workpiece Asset ID} & \textbf{Complementary Fixtures} \\ \hline
\textbf{Workspace}                   & \texttt{conveyor\_belt}, \texttt{conveyor\_ghost}     \\ \hline
\textbf{Model Figurines}             & \texttt{mario}, \texttt{grimlock}, \texttt{bumblebee}, & \texttt{toy\_tray}, \\
\textbf{\& Vehicles}                 & \texttt{dragon}, \texttt{mini\_fire\_engine},          & \texttt{*\_flat\_groove}, \texttt{*\_upright\_groove}, \\
                                     & \texttt{mini\_roller}, \texttt{school\_bus},          & \texttt{*\_v\_groove}, \texttt{*\_flat\_inlay} \\
                                     & \texttt{thomas}, \texttt{porter}                      &                                              \\ \hline
\textbf{Personal Care}               & \texttt{cream\_bottle}, \texttt{cream\_jar},           & \texttt{cosmetics\_tray}, \\
\textbf{\& Cosmetics}                & \texttt{pump\_bottle}, \texttt{serum\_bottle}         & \texttt{*\_flat\_groove}, \texttt{*\_upright\_groove}, \\
                                     &                                                       & \texttt{*\_flat\_v\_groove}, \texttt{*\_v\_groove}, \\
                                     &                                                       & \texttt{*\_flat\_inlay}, \texttt{*\_upright\_inlay} \\ \hline
\textbf{Consumer}                    & \texttt{mouse\_naga}, \texttt{mouse\_white\_taipan},  & \texttt{electronics\_tray}, \\
\textbf{Electronics}                 & \texttt{mouse\_black\_taipan}, \texttt{headset},      & \texttt{*\_upright\_groove}, \texttt{*\_v\_groove}, \\
                                     & \texttt{keyboard}                                     & \texttt{*\_upright\_inlay} \\ \hline
\end{tabularx}
\par\smallskip
\parbox{\linewidth}{\footnotesize * Denotes an object-specific fixture where \texttt{*} maps to the corresponding workpiece Asset ID.}
\end{center}
\end{table}

\subsection{Catalog Composition and Spatial Distribution}
\label{app:catalog_distribution}

\begin{wrapfigure}{r}{0.58\linewidth}
    \centering
    \includegraphics[width=\linewidth]{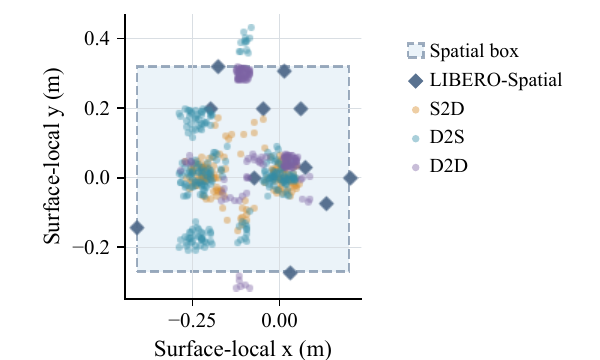}
    \caption{Declared placement-region centers. The dashed LIBERO-Spatial center range contains 510 of 528 Industry region declarations.}
    \label{fig:spatial-region-overlay}
\end{wrapfigure}

The BDDL catalog specifies target workpieces, non-target objects, and placement regions for each suite and level. Figure~\ref{fig:spatial-region-overlay} compares the centers of its declared placement regions with LIBERO-Spatial; each point represents a region declaration, not a task or recorded demonstration.

\paragraph{Target and confounder composition.}
\textbf{D2D.} Level~1 contains four targets without confounders. Level~2 samples two or three targets from four source objects, leaving one or two source confounders. Level~3 contains two or three targets and three confounders distributed across source and destination locations.

\textbf{D2S.} Level~1 contains three moving targets. Level~2 contains three targets and one source confounder. Level~3 contains two moving targets, one or two source confounders, and one prealigned destination object.

\textbf{S2D.} Level~1 contains targets without confounders. Level~2 contains three targets with one or two source confounders. Level~3 contains three targets with upright/flat pose variation and up to three source and destination confounders in total.

\subsection{Language Instruction Construction and Perturbation}
\label{app:language_prompts}

Each task in \liberoindustry is associated with a fixed English instruction that serves as the language-conditioning input to the policy. The instructions were prepared offline through an LLM-assisted authoring workflow using structured task metadata, including the task suite, original manipulation objective, target objects, distractor objects, destination regions, and relevant intrinsic object properties. Controlled language perturbations were then introduced according to three level-specific rules while preserving the task-defining action, target, and destination.

\begin{table}[H]
    \centering
    \small
    \caption{Language-perturbation rules and representative instructions.}
    \label{tab:language_perturbations}
    \begin{tabularx}{\linewidth}{@{}lXX@{}}
        \toprule
        \textbf{Level} & \textbf{Language-perturbation rule} & \textbf{Representative instruction} \\
        \midrule
        LV1 & Lexical and syntactic paraphrasing of the original instruction using common synonyms. & ``Place the pump bottle onto the moving tray.'' \\
        LV2 & Replacement of explicit object names with intrinsic visual descriptions that distinguish the target from distractors. & ``Pick up the semi-transparent bottles with white pumps and place them into the moving conveyor slots.'' \\
        LV3 & Reformulation using industrial process and material-routing terminology while retaining the information required to identify the target and destination. & ``Route all white ambidextrous computer mice from the source conveyor into the moving destination slots.'' \\
        \bottomrule
    \end{tabularx}
\end{table}

Each finalized instruction is stored in \texttt{libero\_suite\_language\_map.py} as a fixed mapping from BDDL task identifiers to instructions and is retrieved deterministically at runtime.

\subsection{Temporal Protocol and Operating Conditions}
\label{app:timing_protocol}

\subsubsection{Terminology and Scope}

Takt time is reserved for a demand-pacing requirement,
\begin{equation}
\tau_{\mathrm{takt}}=
\frac{T_{\mathrm{available}}}{Q_{\mathrm{demand}}},
\end{equation}
which relates available production time to required output
\citep{linck1999takt}. Because the benchmark does not prescribe a customer
demand scenario, its temporal settings are reported as arrival intervals,
service-opportunity intervals, evaluation deadlines, and episode horizons.
These quantities are not interchangeable:
an episode horizon limits recording or rollout duration, whereas an arrival
interval or service-opportunity interval specifies the workload presented to
the controller.

\begin{table}[ht]
\centering
\caption{Temporal quantities used in \liberoindustry.}
\label{tab:timing_terms}
\small
\begin{tabularx}{\linewidth}{@{}l>{\raggedright\arraybackslash}p{0.22\linewidth}X@{}}
\toprule
\textbf{Symbol} & \textbf{Name} & \textbf{Operational meaning} \\
\midrule
$\tau_{\mathrm{takt}}$ & Demand takt & Available production time per required product; used only when demand is specified. \\
$\tau_{\mathrm{src}}$ & Source arrival interval & Time between consecutive source workpieces becoming available. \\
$\tau_{\mathrm{dst}}$ & Destination interval & Time between consecutive destination service opportunities on a pulse-indexed line. \\
$\tau_d$ & Evaluation deadline & Latest admissible completion time for a scored workpiece. \\
$T_{\mathrm{roll}}$ & Episode horizon & Maximum simulated or recorded duration; not a production-rate claim. \\
\bottomrule
\end{tabularx}
\end{table}

\subsubsection{Suite-Specific Physical Timing}

For a continuous source conveyor with signed velocity $V_s$ and consecutive
workpiece pitch $d_{\mathrm{src}}$, the source arrival interval is
\begin{equation}
\tau_{\mathrm{src}}=\frac{d_{\mathrm{src}}}{|V_s|}.
\label{eq:app_source_interval}
\end{equation}
Equivalently, when an arrival interval is authored directly in the task
definition, the corresponding pitch is $d_{\mathrm{src}}=|V_s|\tau_{\mathrm{src}}$.
For a pulse-indexed destination conveyor, let $T_{\mathrm{dwell}}$ be the
stationary service phase and $s_{\mathrm{slot}}$ the distance advanced by one
slot. Its service-opportunity interval is
\begin{equation}
\tau_{\mathrm{dst}}=T_{\mathrm{dwell}}+
\frac{s_{\mathrm{slot}}}{|V_d|},
\label{eq:app_slot_cycle}
\end{equation}
where $V_d$ is the signed destination transfer velocity. The implementation
derives $s_{\mathrm{slot}}$ from the fixture extent along the travel axis plus
the declared inter-fixture clearance. The evaluated release records freeze
the fixture asset, travel axis, clearance, $V_d$, and $T_{\mathrm{dwell}}$ for
every task. S2D timing is therefore determined by the destination mechanism,
not by an assumed robot movement time.

D2D tasks retain both source and destination constraints as the ordered pair
\begin{equation}
\boldsymbol{\tau}_i=
(\tau_{\mathrm{src},i},\tau_{\mathrm{dst},i}),
\label{eq:app_d2d_pair}
\end{equation}
rather than collapsing two asynchronous lines into a single scalar interval.
Table~\ref{tab:suite_timing_sources} summarizes the parameters that must be
exported with the benchmark release.

\begin{table}[ht]
\centering
\caption{Timing provenance by task suite.}
\label{tab:suite_timing_sources}
\small
\begin{tabularx}{\linewidth}{@{}lXXX@{}}
\toprule
\textbf{Suite} & \textbf{Source schedule} & \textbf{Destination schedule} & \textbf{Reported physical limits} \\
\midrule
D2S & $V_s$, $d_{\mathrm{src}}$, $\tau_{\mathrm{src}}$, $t_{\mathrm{off}}$ & Static & $\tau_{\mathrm{src}}$ \\
S2D & Static & $V_d$, $s_{\mathrm{slot}}$, $T_{\mathrm{dwell}}$ & $\tau_{\mathrm{dst}}$ \\
D2D & $V_s$, $d_{\mathrm{src}}$, $\tau_{\mathrm{src}}$, $t_{\mathrm{off}}$ & $V_d$, $s_{\mathrm{slot}}$, $T_{\mathrm{dwell}}$ & $\tau_{\mathrm{src}}$, $\tau_{\mathrm{dst}}$ \\
\bottomrule
\end{tabularx}
\end{table}

The task-specific benchmark execution-time allowances are reported in
Table~\ref{tab:suite_timing}.

\begin{table}[t]
    \centering
    \small
    \caption{Suite-specific nominal benchmark takt and destination service-time settings.}
    \label{tab:suite_timing}
    \setlength{\tabcolsep}{5pt}
    \begin{tabular}{lcccc}
        \toprule
        \textbf{Suite--Level} & \textbf{Takt @ V3/M3} & \textbf{Takt @ V5/M5} & \textbf{Takt @ V8/M8} & \textbf{Service Time} \\
        \midrule
        D2D--LV1 & 5.0~s & 6.0~s & 6.0~s & 1.5~s \\
        D2D--LV2 & 5.0~s & 6.0~s & 6.0~s & 2.0~s \\
        D2D--LV3 & 5.0~s & 6.0~s & 6.0~s & 2.5~s \\
        \midrule
        D2S--LV1 & 4.0~s & 4.0~s & 4.0~s & -- \\
        D2S--LV2 & 5.0~s & 6.0~s & 6.0~s & -- \\
        D2S--LV3 & 5.0~s & 6.0~s & 6.0~s & -- \\
        \midrule
        S2D--LV1 & -- & -- & -- & 1.5~s \\
        S2D--LV2 & -- & -- & -- & 2.0~s \\
        S2D--LV3 & -- & -- & -- & 2.5~s \\
        \bottomrule
    \end{tabular}
    \par\smallskip
    \parbox{\linewidth}{\footnotesize
    \textit{Note.} V3/M3, V5/M5, and V8/M8 denote paired positive and
    negative source-conveyor speeds of $\pm0.03$, $\pm0.05$, and
    $\pm0.08$~m/s, respectively. Each table entry reports the nominal
    benchmark takt assigned at that speed condition.}
\end{table}

\subsubsection{Suite-Specific Episode Horizons}

Equation~\eqref{eq:episode_horizon} expands according to the transport events
encoded by each suite. Each expansion reports both the continuous rollout
duration and the corresponding integer number of control steps.

\paragraph{D2S.}
The horizon covers the source offset, all budgeted source-arrival intervals,
one robot-cycle allowance, and the terminal buffer:
\begin{equation}
\begin{aligned}
T_{\mathrm{roll}}^{\mathrm{D2S}}
  &=t_{\mathrm{off}}+(N_s-1)\tau_{\mathrm{src}}+t_r+t_b,\\
K_{\mathrm{roll}}^{\mathrm{D2S}}
  &=\left\lceil f_cT_{\mathrm{roll}}^{\mathrm{D2S}}\right\rceil .
\end{aligned}
\label{eq:app_d2s_horizon}
\end{equation}

\paragraph{S2D.}
The destination schedule must accommodate the slower of the robot cycle and
the destination service interval for every target:
\begin{equation}
\begin{aligned}
T_{\mathrm{roll}}^{\mathrm{S2D}}
  &=t_i+N_t\max(t_r,\tau_{\mathrm{dst}})+t_b,\\
K_{\mathrm{roll}}^{\mathrm{S2D}}
  &=\left\lceil f_cT_{\mathrm{roll}}^{\mathrm{S2D}}\right\rceil .
\end{aligned}
\label{eq:app_s2d_horizon}
\end{equation}

\paragraph{D2D.}
The horizon combines the source-arrival schedule with one post-arrival
allowance, one destination service interval, and the terminal buffer:
\begin{equation}
\begin{aligned}
T_{\mathrm{roll}}^{\mathrm{D2D}}
  &=t_{\mathrm{off}}+(N_s-1)\tau_{\mathrm{src}}+t_x+\tau_{\mathrm{dst}}+t_b,\\
K_{\mathrm{roll}}^{\mathrm{D2D}}
  &=\left\lceil f_cT_{\mathrm{roll}}^{\mathrm{D2D}}\right\rceil .
\end{aligned}
\label{eq:app_d2d_horizon}
\end{equation}

Here, $N_s$ is the number of budgeted source positions; the generator uses the
four-position source capacity for D2S and D2D Levels~2--3. $N_t$ is the number
of S2D targets. The BDDL value $t_{\mathrm{off}}$ is the time before the first
source arrival, not a spatial offset. During initialization it is converted to
an upstream displacement from the conveyor center as
$d_{\mathrm{off}}=|V_s|t_{\mathrm{off}}$. The fixed values are
$t_{\mathrm{off}}=1.5\,\mathrm{s}$, $t_i=1.0\,\mathrm{s}$,
$t_r=3.0\,\mathrm{s}$, $t_x=1.5\,\mathrm{s}$,
$t_b=1.5\,\mathrm{s}$, and $f_c=20\,\mathrm{Hz}$. The ceiling operation ensures
that the discrete episode includes the complete continuous-time horizon.

\subsubsection{Evaluation Freeze and Throughput Reporting}

The comparison protocol freezes the task definition, timing condition,
deadline grid, episode horizon, control frequency, random seeds, settling
dwell, and success predicate before comparing policies. No timing value is
retuned for an individual policy. Results report the complete
success--deadline curve for each task, alongside the physical
feasibility flags.

The theoretical opportunity rate associated with a scalar interval $\tau$ is
\begin{equation}
q_{\mathrm{opportunity}}=\frac{3600}{\tau}
\quad\text{opportunities/hour}.
\end{equation}
Consequently, intervals of $4$, $5$, and $6$ s correspond to $900$, $720$,
and $600$ theoretical opportunities/hour, respectively. They are not
good-part throughput. Observed good-part throughput is instead
\begin{equation}
q_{\mathrm{good}}=3600\,
\frac{N_{\mathrm{successful}}}{T_{\mathrm{observed}}}
\quad\text{products/hour},
\end{equation}
and is reported as a benchmark-conditioned measurement rather than a universal
HMLV production rate.

\subsection{Conveyor Mechanics and Success Predicates}
\label{app:conveyor_mechanics}

\subsubsection{Simulation Mechanics and Transport Physics}
\label{app:simulation_mechanics}

The simulated transport system maintains physical continuity by updating the state vectors of ungrasped workpieces at every simulation step. Let $\mathbf{R}_B \in \mathbb{SO}(3)$ represent the rotation matrix of the conveyor mechanism in the world coordinate frame, $\hat{\mathbf{a}}_{\text{local}}$ the localized linear travel axis, and $v_{\text{belt}}$ the scalar speed parameter. The instantaneous linear velocity update vector $\mathbf{\dot{q}}_{O_{xy}}(t)$ applied to the horizontal plane of workpiece $O$ is formulated as:
\begin{equation}
    \mathbf{\dot{q}}_{O_{xy}}(t) = \begin{cases} 
        \mathbf{R}_B \cdot \hat{\mathbf{a}}_{\text{local}} \cdot v_{\text{belt}} & \text{if } \Phi_{\text{belt}}(O) \land \neg \Psi_{\text{grasp}}(O) \\ 
        \mathbf{\dot{q}}_{O_{xy}}(t) & \text{otherwise} 
    \end{cases}
\end{equation}
where the Boolean predicate $\Phi_{\text{belt}}(O)$ indicates verified topological contact with the active conveyor mesh, and $\Psi_{\text{grasp}}(O)$ evaluates whether the robot's parallel jaw actuators have established an active, non-slip force closure on the workpiece geometry. Upon satisfying $\Phi_{\text{belt}}(O)$, an instant kinematic anchoring constraint synchronizes the object's base linear velocity directly to the transport vector, preventing non-linear drifting behavior outside explicit physics-driven slippage events.

\subsubsection{Kinematic Overrides for Conveyor and Assembly Physics}
\label{app:kinematic_physics}

To bypass simulation chattering and enforce high-rigidity industrial constraints, we apply explicit position and velocity overrides to dynamic objects at the MuJoCo runtime level. In discrete-time physics engines, relying solely on surface friction to transport complex-mesh workpieces at high velocities can cause unphysical micro-slippage and geometric interpenetration. 

To address these limitations, Algorithm \ref{alg:conveyor_start} dictates the initial conveyor transport kinematics. Specifically, the initial phase utilizes an instant kinematic anchoring mechanism to synchronize the workpiece with the belt upon establishing continuous geometric contact.

\begin{algorithm}[H]
\caption{Conveyor Transport Kinematics (Starting Phase)}
\label{alg:conveyor_start}
\begin{algorithmic}[1]
\Require $\text{sim.data}$, $\text{target\_vel}$, $\text{obj\_name}$, $\text{belt\_geom\_ids}$
\State $\text{is\_contact} \gets \text{\_in\_contact\_with}(\text{sim}, \text{obj\_geom\_ids}, \text{belt\_geom\_ids})$
\If{not $\text{is\_contact}$}
    \State \Return \Comment{Allow standard gravity and free rigid-body dynamics}
\EndIf
\State $\text{qpos\_adr}, \text{qvel\_adr} \gets \text{GetJointAddresses}(\text{sim}, \text{obj\_body\_id})$
\If{$\text{obj\_name} \notin \text{\_anchored\_quats}$}
    \State $\text{\_anchored\_quats}[\text{obj\_name}] \gets [1.0, 0.0, 0.0, 0.0]$
    \State $\text{\_anchored\_pos}[\text{obj\_name}] \gets \text{sim.data.qpos}[\text{qpos\_adr}:\text{qpos\_adr}+2]$
\EndIf
\State \Comment{Enforce orientation lock to prevent mesh interpenetration jitter}
\State $\text{sim.data.qpos}[\text{qpos\_adr}+3:\text{qpos\_adr}+7] \gets \text{\_anchored\_quats}[\text{obj\_name}]$
\State \Comment{Override velocities based on belt target speed}
\For{$i \in \{0, 1\}$}
    \If{$|\text{target\_vel}[i]| < 1\text{e-}5$}
        \State $\text{sim.data.qpos}[\text{qpos\_adr}+i] \gets \text{\_anchored\_pos}[\text{obj\_name}][i]$
    \EndIf
\EndFor
\end{algorithmic}
\end{algorithm}

\vspace{1em}

Following the initial transport phase, Algorithm \ref{alg:slot_ending} governs the terminal insertion constraints into the moving conveyor slots. Subsequently, this terminal phase employs a multi-stage velocity arrest mechanism (Levels 1 to 3) to lock the workpiece inside the slot geometry upon satisfying strict dot-product orientation thresholds.

\begin{algorithm}[H]
\caption{Multi-Stage Slot Arrest Kinematics (Ending Target)}
\label{alg:slot_ending}
\begin{algorithmic}[1]
\Require $\text{sim.data}$, $\text{vel\_vec}$, $\text{delta\_pos}$, $\text{lock\_level}$
\State $\text{obj\_qpos\_adr}, \text{obj\_qvel\_adr} \gets \text{GetJointAddresses}(\text{sim}, \text{obj\_body\_id})$

\If{$\text{AlignmentThresholdMet}()$}
    \If{$\text{lock\_level} == 1$}
        \State \Comment{Level 1: Planar Lock Only (X, Y locked to conveyor; Z/Rotation free)}
        \State $\text{sim.data.qvel}[\text{obj\_qvel\_adr}:\text{obj\_qvel\_adr}+2] \gets \text{vel\_vec}[0:2]$
        \State $\text{sim.data.qpos}[\text{obj\_qpos\_adr}:\text{obj\_qpos\_adr}+2] \mathrel{+}= \text{delta\_pos}[0:2]$
    \Else
        \State \Comment{Level 2 \& 3: Full 3D Spatial Lock}
        \State $\text{sim.data.qvel}[\text{obj\_qvel\_adr}:\text{obj\_qvel\_adr}+3] \gets \text{vel\_vec}$
        \State $\text{sim.data.qpos}[\text{obj\_qpos\_adr}:\text{obj\_qpos\_adr}+3] \mathrel{+}= \text{delta\_pos}$
        \State $\text{sim.data.qvel}[\text{obj\_qvel\_adr}+2] \gets 0.0$ \Comment{Arrest Z-axis gravity}
    \EndIf
\EndIf
\State \Comment{Update moving groove target position continually}
\State $\text{sim.data.qpos}[\text{qpos\_adr}:\text{qpos\_adr}+3] \gets \text{exact\_world\_pos}$
\State $\text{sim.data.qpos}[\text{qpos\_adr}+3:\text{qpos\_adr}+7] \gets \text{world\_groove\_quat\_wxyz}$
\end{algorithmic}
\end{algorithm}

\subsubsection{Predicate Scaling and Constraint Matrix}
\label{app:kinematic_predicates}

To accommodate the progressive complexity of industrial workflows, the state evaluation predicates are stratified across three difficulty levels. As environmental variance increases, the task definitions impose stricter kinematic constraints and require higher degrees of freedom (DoF) for successful execution.

\begin{table}[htbp]
    \centering
    \caption{Kinematic Constraint Matrix by Predicate. A checkmark (\checkmark) indicates the mechanism is strictly verified for task success, while a cross ($\times$) indicates it is relaxed or not applicable.}
    \label{tab:predicate_mechanisms}
    \resizebox{\linewidth}{!}{%
    \begin{tabular}{@{}lcccc@{}}
        \toprule
        \textbf{Predicate} & \textbf{Gripper Released} & \textbf{Topological Contact} & \textbf{Vector Axis Alignment} & \textbf{Velocity Sync} \\
        \midrule
        \texttt{ExactOn}   & $\checkmark$ & $\checkmark$ & $\times$     & $\times$     \\
        \texttt{DynamicOn} & $\checkmark$ & $\checkmark$ & $\times$     & $\checkmark$ \\
        \texttt{AlignedOn} & $\checkmark$ & $\checkmark$ & $\checkmark$ & $\times$     \\
        \texttt{Routed}    & $\checkmark$ & $\checkmark$ & $\checkmark$ & $\checkmark$ \\
        \bottomrule
    \end{tabular}}
\end{table}

The mapping of these target definition predicates across task paradigms is summarized in Table~\ref{tab:predicate_matrix}.

\begin{table}[htbp]
    \centering
    \caption{Target Definition Predicates and Kinematic Constraints by Difficulty Level.}
    \label{tab:predicate_matrix}
    \resizebox{\linewidth}{!}{%
    \begin{tabular}{@{}llccc@{}}
        \toprule
        \textbf{Difficulty Level} & \textbf{Receptacle Type} & \textbf{Static Target (D2S)} & \textbf{Dynamic Target (S2D, D2D)} & \textbf{Control Constraint} \\
        \midrule
        L1  & Tray            & \texttt{ExactOn}   & \texttt{DynamicOn} & 3-DoF \\
        L2  & Slot / V-Groove & \texttt{AlignedOn} & \texttt{Routed}    & 5-DoF \\
        L3  & Inlay          & \texttt{AlignedOn} & \texttt{Routed}    & 6-DoF \\
        \bottomrule
    \end{tabular}}
\end{table}

\subsection{Environmental Generalization Settings}
\label{app:environmental_specs}

\subsubsection{Standards and Reporting Basis}

The perturbation families are organized using selected industrial measurement and operating terminology. ISO~9283 provides the pose-accuracy and pose-repeatability framework used to describe bounded object, fixture, and camera-pose deviations \citep{iso9283_1998}. ISO~8995-1 provides the workplace-lighting context for controlled illumination changes \citep{iso8995_2002}, while GB/T~10595 provides the operating context for the commanded belt-conveyor conditions \citep{gbt10595_2017}. These standards provide reporting context but neither prescribe nor validate the simulator-domain bounds below, which are fixed by the benchmark generator.

\subsubsection{Perturbation Specifications}

Level~1 is the nominal LIBERO-Industry condition. Levels~2 and~3 introduce moderate and broad generalization conditions, respectively.

\paragraph{Workspace Geometry.}

For object initialization, Level~1 scales each declared planar placement interval by $0.5$ and fixes yaw. Levels~2 and~3 scale the interval by $0.2$ and sample yaw from $\mathcal{U}(0,2\pi)$. D2D fixture translations are bounded by $\pm2$ cm in Levels~1 and~2; Level~2 additionally applies a shared yaw perturbation of $\pm5^\circ$. Level~3 applies cardinal transpositions, bounds translation to $\pm2$ cm in $x$ and to $\pm2$ or $\pm5$ cm in $y$ according to the transposition, and applies yaw perturbations of $\pm5^\circ$ or $\pm10^\circ$.

D2S Level~1 uses common fixture translations of up to $\pm5$ cm. Level~2 retains geometry-bounded translations of up to $\pm5$ cm and applies belt and groove yaw perturbations of $\pm10^\circ$. Level~3 combines cardinal transpositions with common translations of up to $\pm2$ cm in $x$ and $\pm4$ cm in $y$, $5$ mm local inlay jitter, and yaw perturbations of $\pm10^\circ$. S2D uses common translations of up to $\pm5$ cm in all tiers, adds independent source and destination yaw perturbations of $\pm10^\circ$ at Level~2, and combines cardinal transpositions with $\pm20^\circ$ yaw perturbations at Level~3.

\paragraph{Visual Observations.}

Table~\ref{tab:app_shared_perturbations} reports the camera, illumination, and workbench parameters shared across D2D, D2S, and S2D tasks.

\begin{table*}[t]
    \centering
    \caption{Visual-observation perturbations across the three generalization tiers. Angular bounds are in degrees; translation bounds are Euclidean norms.}
    \label{tab:app_shared_perturbations}
    \scriptsize
    \setlength{\tabcolsep}{3pt}
    \renewcommand{\arraystretch}{1.12}
    \begin{tabularx}{\textwidth}{@{}p{0.18\textwidth}XXX@{}}
        \toprule
        \textbf{Configuration} & \textbf{Level 1: nominal} & \textbf{Level 2: moderate} & \textbf{Level 3: broad} \\
        \midrule
        Agent-view orbit & Azimuth $0$; elevation $38.922340$; distance $1.130651$ m & Azimuth $\{45,315\}\pm5$; elevation $38.922340\pm5$; distance $1.130651\pm0.1$ m & Azimuth sampled from $[0,135)\cup(230,360)$; elevation $38.922340\pm10$; distance $1.130651\pm0.2$ m \\
        Agent-view rotation & Fixed & Rotation norm $\leq1$ & Rotation norm $\leq2$ \\
        Wrist camera & Not used & Translation norm $\leq2.29$ mm; rotation norm $\leq1$ & Translation norm $\leq4.57$ mm; rotation norm $\leq2$ \\
        Illumination & Neutral profile; scale $2.5$; no cast shadows & Neutral, warm, or cool profile; scale $2.5\pm0.5$; cast shadows & Neutral, warm, or cool profile; scale $2.5\pm1.0$; cast shadows \\
        Workbench surface & Green matte & Green, blue, or black; matte or glossy & Same six controlled styles as Level~2 \\
        \bottomrule
    \end{tabularx}
\end{table*}

The illumination entries are normalized MuJoCo renderer parameters rather than photometric measurements in lux. They therefore provide reproducible simulator-domain conditions without asserting photometric conformity.

\paragraph{Transport Dynamics.}

The named conveyor-speed conditions are anchored at $0.03$, $0.05$, and $0.08$ m/s. D2D samples magnitudes within three non-overlapping intervals: $[0.0300,0.0399]$, \allowbreak $[0.0400,0.0649]$, and \allowbreak $[0.0650,0.0800]$ m/s. Level~1 uses equal signed source and destination speeds, Level~2 uses unequal same-direction speeds, and Level~3 uses opposite directions. D2D service times are $1.5$, $2.0$, and $2.5$ s for Levels~1--3.

For D2S, Level~1 applies $\pm5\%$ relative speed variation around the selected condition, Level~2 applies $\pm0.005$ m/s absolute variation, and Level~3 uses the selected condition without additional speed noise. S2D uses the same Level~1 and Level~2 rules; Level~3 samples speed magnitude uniformly from $[0.03,0.08]$ m/s while retaining the task direction. S2D service times are $1.5$, $2.0$, and $2.5$ s for Levels~1--3.

\subsubsection{Reproducibility and Configuration}

All scene perturbations are deterministic for the task seed. For each task, the generator serializes the camera, lighting, surface, reset, and rollout parameters. The record is stored in \path{libero/libero/benchmark/industry_env_args.json}. Suite-specific geometry, composition, and transport rules are defined in \path{notebooks/industry/autogen/configs.py}, \path{perturbations.py}, \path{task_composition.py}, and the D2D, D2S, and S2D generation scripts.


\section{Evaluation Protocol and Extended Results}
\label{app:evaluation_details}

\subsection{Evaluated Models and Checkpoints}
\label{app:model_details}

We evaluate five publicly released policies fine-tuned on the original LIBERO
benchmark, without further fine-tuning on \liberoindustry. This is a
cross-benchmark transfer evaluation, not an evaluation of unfine-tuned base
models. Table~\ref{tb:model_access} identifies the checkpoint sources.
The exact weight revisions and model-specific inference settings should be
recorded in the released evaluation configurations.

\begin{table}[htbp]
\caption{Original-LIBERO-adapted checkpoints used for cross-benchmark evaluation.}
\label{tb:model_access}
\centering
\footnotesize
\setlength{\tabcolsep}{5pt}
\renewcommand{\arraystretch}{1.18}
\begin{tabularx}{\linewidth}{@{}>{\raggedright\arraybackslash}p{0.22\linewidth}>{\raggedright\arraybackslash}X@{}}
\toprule
\textbf{Policy} & \textbf{Checkpoint identifier} \\
\midrule
$\pi_{0.5}$ & \nolinkurl{gs://openpi-assets/checkpoints/pi05_libero} \\
OpenVLA-OFT & \href{https://huggingface.co/moojink/openvla-7b-oft-finetuned-libero-spatial-object-goal-10}{\nolinkurl{moojink/openvla-7b-oft-finetuned-libero-spatial-object-goal-10}} \\
UniVLA & \href{https://huggingface.co/qwbu/univla-7b-224-sft-libero}{\nolinkurl{qwbu/univla-7b-224-sft-libero}} \\
X-VLA & \href{https://huggingface.co/2toINF/X-VLA-Libero}{\nolinkurl{2toINF/X-VLA-Libero}} \\
Cosmos Policy & \href{https://huggingface.co/nvidia/Cosmos-Policy-LIBERO-Predict2-2B}{\nolinkurl{nvidia/Cosmos-Policy-LIBERO-Predict2-2B}} \\
\bottomrule
\end{tabularx}
\end{table}

\subsection{Evaluation-Set Breakdown}
\label{app:evaluation_set_breakdown}

Table~\ref{tab:initial_slotting_baseline} reports the initial-state slotting baseline, followed by the suite-level results in Table~\ref{tab:main_results_by_suite} for the same five checkpoints as Table~\ref{tab:main_results}.

\begin{table}[!htbp]
\caption{\textbf{BDDL-declared composition of the 36-task held-out evaluation split.} $N_{\mathrm{target}}$ denotes goal workpieces requiring manipulation and excludes initially valid destination workpieces; $N_{\mathrm{base}}$ denotes pre-existing target confounders. Counts are computed from the four selected BDDLs in each suite--level. The final column is the macro-average of task-level ratios, not a ratio of pooled counts.}
\label{tab:initial_slotting_baseline}
\centering
\footnotesize
\setlength{\tabcolsep}{3pt}
\begin{tabular*}{\linewidth}{@{\extracolsep{\fill}}lcccc@{}}
\toprule
\textbf{Suite} & \shortstack{$N_{\mathrm{target}}$\\(L1/L2/L3)} & \shortstack{$N_{\mathrm{base}}$\\(L1/L2/L3)} & \shortstack{Count\\$N_{\mathrm{base}}/(N_{\mathrm{target}}+N_{\mathrm{base}})$} & \shortstack{$\mathrm{PCR}_{\mathrm{S}}^{\mathrm{init}}$\\(\%)} \\
\midrule
S2D & $12/12/12=36$ & $0/0/6=6$ & $6/42$ & 10.83 \\
D2S & $12/12/8=32$ & $0/0/4=4$ & $4/36$ & 11.11 \\
D2D & $12/9/10=31$ & $0/0/8=8$ & $8/39$ & 15.00 \\
\textbf{Overall} & $36/33/30=99$ & $0/0/18=18$ & $18/117$ & \textbf{12.31} \\
\bottomrule
\end{tabular*}
\end{table}

\begin{table}[!htbp]
\caption{\textbf{Evaluation-set results by suite.} Scores in \%; 120 rollouts per policy per suite.}
\label{tab:main_results_by_suite}
\centering
\footnotesize
\setlength{\tabcolsep}{2pt}
\begin{tabular*}{\linewidth}{@{\extracolsep{\fill}}l*{9}{c}@{}}
\toprule
& \multicolumn{3}{c}{S2D} & \multicolumn{3}{c}{D2S} & \multicolumn{3}{c}{D2D} \\
\cmidrule(lr){2-4}\cmidrule(lr){5-7}\cmidrule(l){8-10}
Policy & SR & $\mathrm{PCR}_{\mathrm{T}}$ & $\mathrm{PCR}_{\mathrm{S}}$ & SR & $\mathrm{PCR}_{\mathrm{T}}$ & $\mathrm{PCR}_{\mathrm{S}}$ & SR & $\mathrm{PCR}_{\mathrm{T}}$ & $\mathrm{PCR}_{\mathrm{S}}$ \\
\midrule
$\pi_{0.5}$ & 0.0 & 1.1 & 11.9 & 0.0 & 0.0 & 11.1 & 0.0 & 0.0 & 14.3 \\
OpenVLA-OFT & 0.0 & 0.0 & 9.1 & 0.0 & 0.0 & 11.0 & 0.0 & 0.0 & 13.4 \\
UniVLA & 0.0 & 0.0 & 10.8 & 0.0 & 0.0 & 11.1 & 0.0 & 0.0 & 14.7 \\
X-VLA & 0.0 & 0.0 & 10.8 & 0.0 & 0.0 & 11.1 & 0.0 & 0.0 & 13.8 \\
Cosmos Policy & 0.0 & 0.0 & 10.8 & 0.0 & 0.0 & 10.0 & 0.0 & 0.0 & 14.6 \\
\bottomrule
\end{tabular*}\par\smallskip
{\scriptsize\raggedright $\mathrm{PCR}_{\mathrm{S}}$ includes initially valid destination workpieces (Section~\ref{sec:evaluation_protocol}).\par}
\end{table}

\subsection{Cumulative Diagnostic Stress Progression}
\label{app:diagnostic_stress_progression}

\subsubsection{Stress-Progression Setup}
\label{app:diagnostic_stress_setup}

Unless otherwise noted, the diagnostic progression uses the same evaluation setup as Section~\ref{subsec:experimental_setup}.
Table~\ref{tab:diagnostic_stress_setup} records the progression-specific task coverage, rollout counts, and language conditions.

\begin{table}[H]
\caption{Task and language composition of the cumulative diagnostic stress progression.}
\label{tab:diagnostic_stress_setup}
\centering
\footnotesize
\setlength{\tabcolsep}{4pt}
\renewcommand{\arraystretch}{1.15}
\begin{tabularx}{\linewidth}{@{}l l c X@{}}
\toprule
\textbf{Stage} & \textbf{Registered tasks} & \textbf{Trials} & \textbf{Language condition} \\
\midrule
D0 & ABLD000--009 & $10\times20=200$ & Ten original spatial-relation prompts, such as ``between the plate and the ramekin.'' \\
D1 & ABLD100--109 & $10\times20=200$ & Generic prompt: ``pick up the akita black bowl and place it on the plate.'' \\
D2 & ABLD200--211 & $12\times20=240$ & Three singular D2D/D2S/S2D templates without a conveyor-source reference. \\
D3 & ABLD300--311 & $12\times20=240$ & Singular family-specific templates aligned with the introduced conveyor geometry. \\
D4 & ABLD400--411 & $12\times20=240$ & D2D and S2D introduce ``moving tray''; D2S wording remains unchanged. \\
D5 & ABLD500--511 & $12\times20=240$ & The D4 templates use plural object nouns. \\
D6 & ABLD600--611 & $12\times20=240$ & The plural D5 instruction templates are retained. \\
\bottomrule
\end{tabularx}
\end{table}

\subsubsection{Trajectory-Level Failure Classification}
\label{app:failure_attribution}

\begin{figure}[H]
    \centering
    \includegraphics[width=0.98\textwidth]{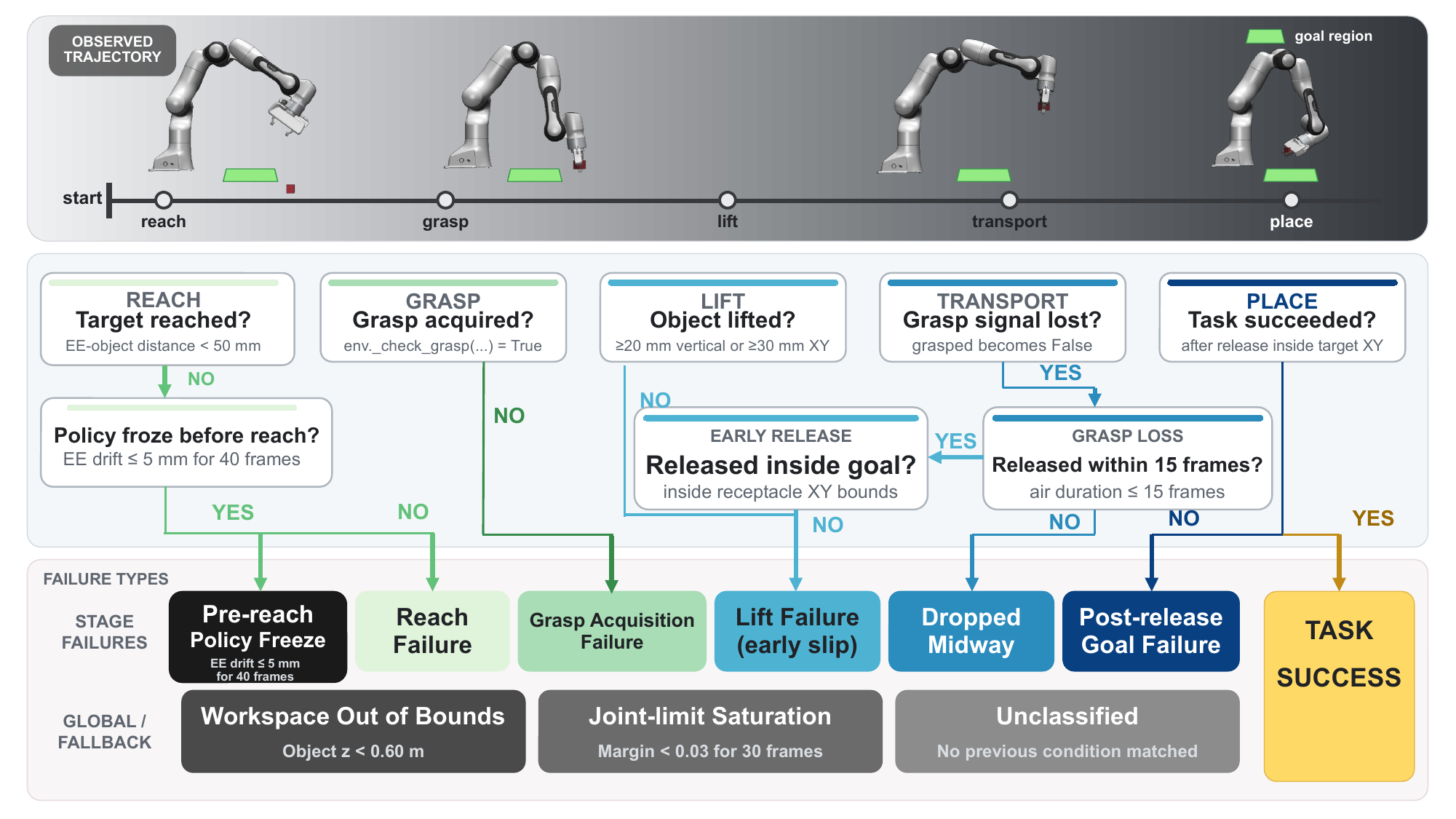}
    \caption{%
        \textbf{Trajectory-level failure classification.}
        Each episode is evaluated along the reach--grasp--lift--transport--place sequence.
        The decision rules expose the operational thresholds used for stage-specific labels, while workspace violations, joint-limit saturation, and unmatched cases remain global or fallback outcomes.
    }
    \label{fig:failure-tracker}
\end{figure}

Figure~\ref{fig:failure-tracker} maps each rollout to the first unmet manipulation phase, from pre-reach policy freeze and reach failure through post-release goal failure.

The resulting labels are used in a cumulative diagnostic stress progression, not a one-factor-at-a-time causal ablation.
From D0 to D6, the sequence adds an industrial background, removes a visual anchor, replaces the target with industrial trays, introduces a conveyor and temporal dynamics, and increases the sequence from one to three objects.
Adjacent stages may change more than one factor, so their differences localize breakdowns without isolating causes.
The D0--D6 results aggregate 200 rollouts at D0--D1 and 240 rollouts at D2--D6, with 20 indexed episodes per registered task. Numeric evaluator seeds were not retained; accordingly, we report the observed proportions descriptively and do not claim a prespecified confidence-interval analysis.
D6 has no successful trajectory, so its policy-step statistic is undefined.

\subsubsection{Representative Failure Trajectories}
\label{app:representative_failure_trajectories}

\begin{figure}[!htbp]
    \centering
    \footnotesize
    \setlength{\tabcolsep}{3pt}
    \renewcommand{\arraystretch}{1.08}
    \begin{tabularx}{\textwidth}{@{}p{0.14\textwidth} p{0.17\textwidth} p{0.17\textwidth} p{0.17\textwidth} X@{}}
        \toprule
        \textbf{Failure phase} & \multicolumn{3}{c}{\textbf{Representative trajectory frames}} & \textbf{Tracker outcome} \\
        \cmidrule(lr){2-4}
        & \multicolumn{1}{c}{$t_1$} & \multicolumn{1}{c}{$t_2$} & \multicolumn{1}{c}{$t_3$} & \\
        \midrule
        \textbf{Spatial / reach}\par
        {\scriptsize D5; ABLD503; episode 14}
        & {\centering\includegraphics[width=\linewidth]{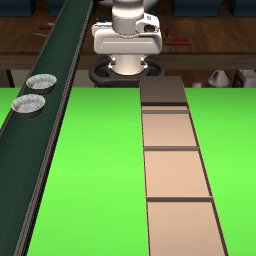}\par\scriptsize frame 0\par}
        & {\centering\includegraphics[width=\linewidth]{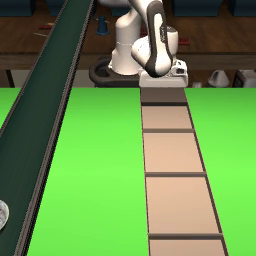}\par\scriptsize frame 189\par}
        & {\centering\includegraphics[width=\linewidth]{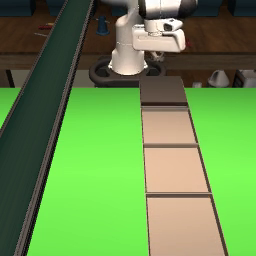}\par\scriptsize frame 399\par}
        & \textbf{Spatial approach failure}\par
        {\scriptsize 132/240 episodes (55.0\%) at D5.} \\
        \midrule
        \textbf{Grasp / lift}\par
        {\scriptsize D3; ABLD306; episode 10}
        & {\centering\includegraphics[width=\linewidth]{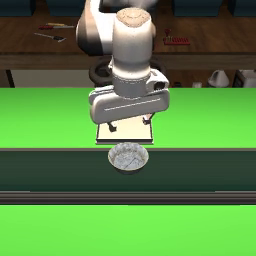}\par\scriptsize frame 42\par}
        & {\centering\includegraphics[width=\linewidth]{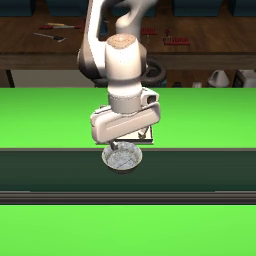}\par\scriptsize frame 147\par}
        & {\centering\includegraphics[width=\linewidth]{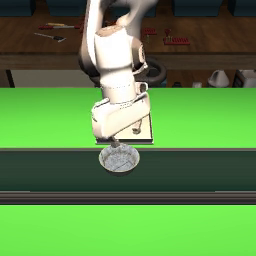}\par\scriptsize frame 399\par}
        & \textbf{Kinematic lift failure}\par
        {\scriptsize 9/240 episodes (3.8\%) at D3.} \\
        \midrule
        \textbf{Dynamic transport}\par
        {\scriptsize D4; ABLD406; episode 11}
        & {\centering\includegraphics[width=\linewidth]{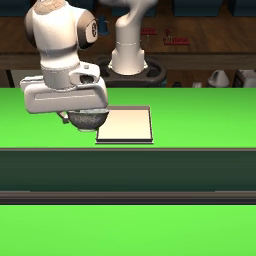}\par\scriptsize frame 84\par}
        & {\centering\includegraphics[width=\linewidth]{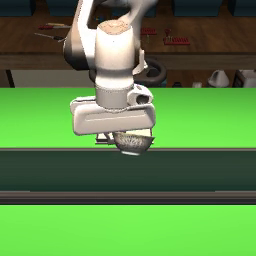}\par\scriptsize frame 168\par}
        & {\centering\includegraphics[width=\linewidth]{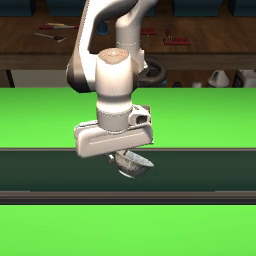}\par\scriptsize frame 231\par}
        & \textbf{Dynamic slip}\par
        {\scriptsize Dropped midway; 35/240 episodes (14.6\%) at D4.} \\
        \midrule
        \textbf{Placement}\par
        {\scriptsize D5; ABLD509; episode 15}
        & {\centering\includegraphics[width=\linewidth]{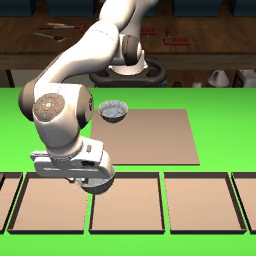}\par\scriptsize frame 160\par}
        & {\centering\includegraphics[width=\linewidth]{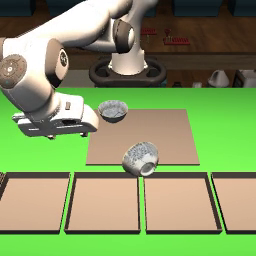}\par\scriptsize frame 200\par}
        & {\centering\includegraphics[width=\linewidth]{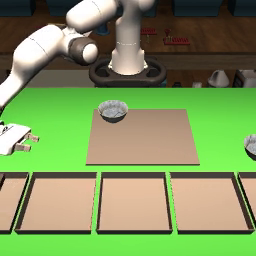}\par\scriptsize frame 399\par}
        & \textbf{Post-release goal failure}\par
        {\scriptsize 28/240 episodes (11.7\%) at D5.} \\
        \bottomrule
    \end{tabularx}
    \caption{%
        \textbf{Representative failure trajectories.}
        Each row shows three temporally ordered frames from one failed rollout, followed by the trajectory-level classifier output and its frequency within the corresponding diagnostic stage.
        The examples link the aggregate categories in Figure~\ref{fig:ablation-progression} to observable trajectory behavior and are illustrative rather than exhaustive.
    }
    \label{fig:representative-failure-trajectories}
\end{figure}

To complement the aggregate failure proportions in Figure~\ref{fig:ablation-progression}, Figure~\ref{fig:representative-failure-trajectories} shows representative frame sequences from four classified rollouts.
Each row contains one trajectory from the frozen $\pi_{0.5}$ tracker run and reports the corresponding stage-level frequency.
The examples illustrate the operational meaning of the tracker outcomes; they are not an exhaustive validation of all classified episodes.

\section{Cross-Embodiment Conversion and Policy Adaptation}
\label{app:cross_embodiment}

\paragraph{Embodiment and task-space tracking.}
We instantiate an AUBO-i5H arm with a CTAG2F90D gripper, using the gripper's
\texttt{grip\_site} as the tool center point (TCP) provided by the factory provider. Panda and AUBO environments
share the task definition and named non-robot scene state, while the AUBO-i5H arm
and controller are initialized natively. After settling, corresponding task-body
poses must agree within $10^{-6}$\,m and $10^{-3}$ degrees. Because the arms have
seven and six joints, respectively, conversion tracks Cartesian references
rather than copying joint coordinates. In the shared simulator world frame,
the AUBO-i5H action is
\begin{equation}
\mathbf{u}^{A}_{t} =
\begin{bmatrix}
\operatorname{clip}_{[-1,1]}\!\left(
\begin{bmatrix}
k(\mathbf{p}^{P}_{t+1}-\mathbf{p}^{A}_{t})/s_p \\
k\operatorname{Log}\!\left(R^{P}_{t+1}(R^{A}_{t})^{\top}\right)/s_R
\end{bmatrix}\right) \\
g^{P}_{t}
\end{bmatrix},
\end{equation}
where $k=3$, $s_p=0.05$\,m, $s_R=0.5$\,rad, and $\operatorname{Log}$ returns a
three-dimensional rotation vector. The target controller follows the next
Panda TCP reference online while retaining the source gripper schedule.
This aligns task-space references, not joint trajectories or pixelwise camera
observations.

\begin{figure}[H]
    \centering
    \setlength{\tabcolsep}{1pt}
    \begin{tabular}{@{}cccc@{}}
        \includegraphics[width=.24\linewidth]{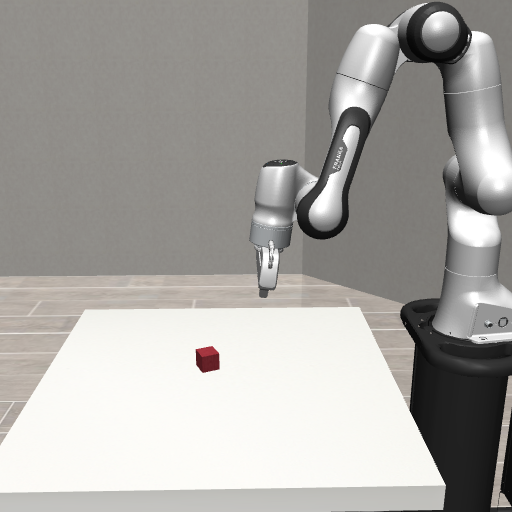} &
        \includegraphics[width=.24\linewidth]{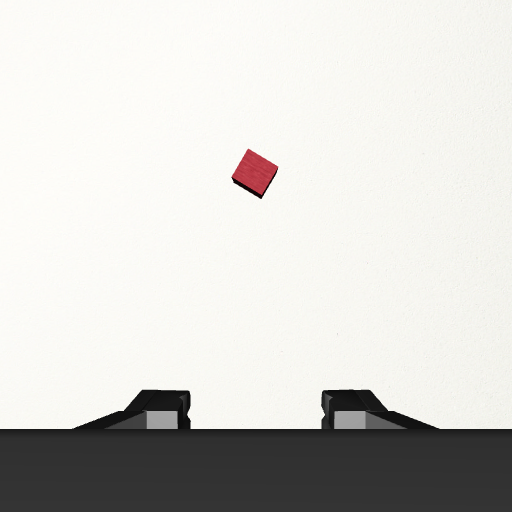} &
        \includegraphics[width=.24\linewidth]{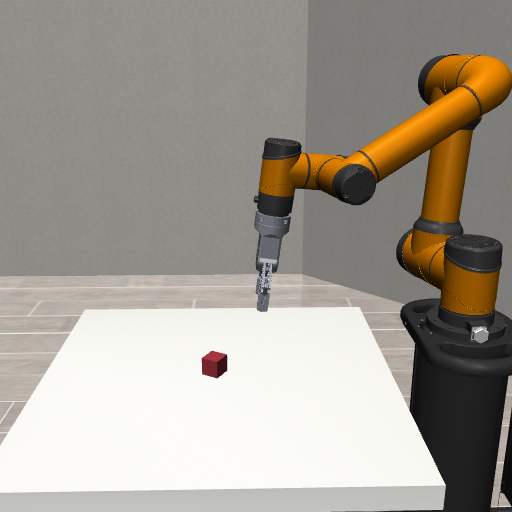} &
        \includegraphics[width=.24\linewidth]{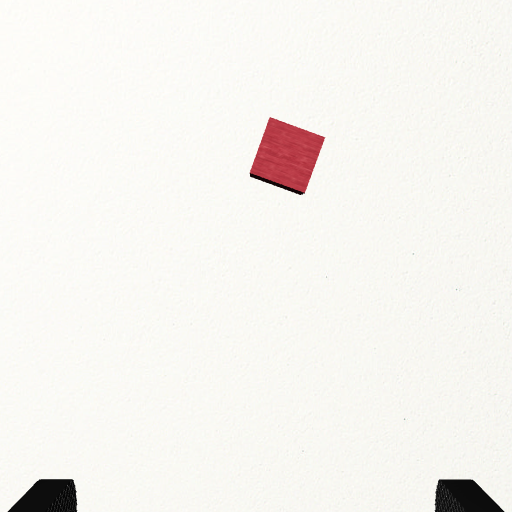} \\
        \scriptsize (a) Panda setup & \scriptsize (b) Panda wrist &
        \scriptsize (c) AUBOi5H--CTAG2F90D setup & \scriptsize (d) AUBO-i5H wrist
    \end{tabular}
    \caption{Representative Lift-scene setup and wrist observations.  The
    comparison checks robot--gripper composition and view coverage; it does not
    assume pixelwise camera equivalence.}
    \label{fig:panda_AUBO_embodiments}
\end{figure}

\paragraph{Policy interface and visual convention.}
The policy receives external and wrist RGB images and a seven-dimensional
state comprising TCP position, axis--angle orientation, and one CTAG driver
coordinate. Its seven-dimensional action comprises six relative
operational-space controller commands and one gripper command; no additional
delta-action transform is applied. Images are stored at $256\times256$ and
resized to $224\times224$ by OpenPI. For the aligned simulation releases,
export and inference both use the same $180$-degree rotation of raw simulator
images. Since the recorded HDF5 images already contain a vertical flip,
export applies a horizontal flip to obtain this convention.

\subsection{Trajectory Conversion and Recovery}
\label{app:mimicgen_recovery}

\paragraph{Motivation and scope.}
Motivated by MimicGen's demonstrated transfer of the same Panda demonstrations
to Sawyer, IIWA, and UR5e~\citep{mandlekar2023mimicgen}, we bridge the
Panda-to-AUBO embodiment gap with a MimicGen-based recovery procedure that
transforms object-centric end-effector segments into AUBO reference frames and
retains only trajectories that pass three out of three replays. Our local
implementation follows MimicGen's subtask decomposition and reference-frame
transformation principles, but uses our own LIBERO and AUBO execution,
validation, and data-materialization code rather than the released software.

\begin{figure}[H]
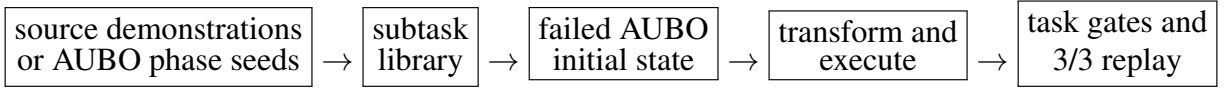

    \centering
    \resizebox{\columnwidth}{!}{%
    \begin{tabular}{c@{\;\(\rightarrow\)\;}c@{\;\(\rightarrow\)\;}c@{\;\(\rightarrow\)\;}c@{\;\(\rightarrow\)\;}c}
    \fbox{\shortstack{source demonstrations\\or AUBO phase seeds}} &
    \fbox{\shortstack{subtask\\library}} &
    \fbox{\shortstack{failed AUBO\\initial state}} &
    \fbox{\shortstack{transform and\\execute}} &
    \fbox{\shortstack{task gates and\\3/3 replay}}
    \end{tabular}}
    \caption{Recovery pipeline. A candidate is added to the training dataset
    only after generation and replay validation.}
    \label{fig:mimicgen_AUBO_recovery}
\end{figure}

\paragraph{Recovery procedure.}
Gripper events and task predicates split successful source rollouts into
object-centric segments. LIBERO-Object and LIBERO-Spatial use separate pick and
place segments from successful Panda demonstrations, whereas LIBERO-Goal and
LIBERO-10 use object, articulated-fixture, and push phases selected from
successful AUBO phase seeds to preserve long-horizon task invariants. Given a
source end-effector pose ${}^{W}\!T^{s}_{e,t}$ and the source and query reference
poses ${}^{W}\!T^{s}_{r}$ and ${}^{W}\!T^{q}_{r}$, respectively, each segment is
transformed as

\begin{equation}
    {}^{W}\!T^{q}_{e,t}
    = {}^{W}\!T^{q}_{r}
      \left({}^{W}\!T^{s}_{r}\right)^{-1}
      {}^{W}\!T^{s}_{e,t},
    \label{eq:mimicgen_AUBO_transform}
\end{equation}

which preserves the end-effector motion relative to the live object or fixture.
Each attempt restores the failed conversion's exact AUBO initial state, bridges
to the transformed segment by Cartesian interpolation, retains the demonstrated
gripper schedule, and executes the poses with the AUBO operational-space
controller. A candidate is retained only if its task predicates hold, its arrays
are finite and actions bounded, its joint-margin and controller-clipping checks
pass, terminal success persists, and its stored actions succeed in three replays
from the same initial state and scene. Accepted recoveries retain their
demonstration identity and provenance and are merged only with replay-verified
raw conversions, as summarized in Table~\ref{tab:mimicgen_recovery_yield}.

\begin{table}[t]
    \centering
    \caption{Final Panda-to-AUBO conversion and recovery yield. Raw retained
    demonstrations passed all three exact-action replays. Recovery includes
    both direct conversion failures and replay-rejected raw demonstrations.}
    \label{tab:mimicgen_recovery_yield}
    \footnotesize
    \resizebox{\linewidth}{!}{%
    \begin{tabular}{lrrrrr}
        \toprule
        Suite &
        Raw 3/3 retained &
        Replay-rejected &
        Direct raw failures &
        Recovered 3/3 &
        Final retained \\
        \midrule
        LIBERO-Object  & 425 & 4  & 71  & 75  & 500 \\
        LIBERO-Spatial & 147 & 38 & 315 & 353 & 500 \\
        LIBERO-Goal    & 216 & 41 & 243 & 284 & 500 \\
        LIBERO-10      & 184 & 37 & 279 & 316 & 500 \\
        \midrule
        Total          & 972 & 120 & 908 & 1,028 & 2,000 \\
        \bottomrule
    \end{tabular}}
\end{table}


\section{Physical System, Real-World Dataset, and Sim--Real Evaluation}
\label{app:real_world_system}

\subsection{Physical Workcell and Sensor Setup}
\label{app:physical_workcell}

The physical dynamic-to-static (D2S) workcell uses an AUBO-i5H arm with a
CTAG2F90D gripper to transfer objects from a moving conveyor into stationary
target fixtures. A Pico headset and controllers support teleoperated
demonstration collection. An external HIKON camera and a wrist camera provide
the two recorded RGB views
(Figure~\ref{fig:rw_physical_setup}); the policy's state and action interface
is specified in Appendix~\ref{app:cross_embodiment}. The target fixture changes
from a tray in Level~1 to object-matched grooves in Level~2 and form-fitting
inlays in Level~3. The robot and two-view policy interface are shared across
levels while placement constraints change at the destination.

\begin{figure}[H]
    \centering
    \begin{minipage}[t]{0.42\linewidth}
        \vspace{0pt}
        \centering
        \begingroup
        \setlength{\unitlength}{\linewidth}
        \begin{picture}(1,1.021)
            \put(0,0){\includegraphics[width=\linewidth,trim=0bp 150bp 0bp 250bp,clip]{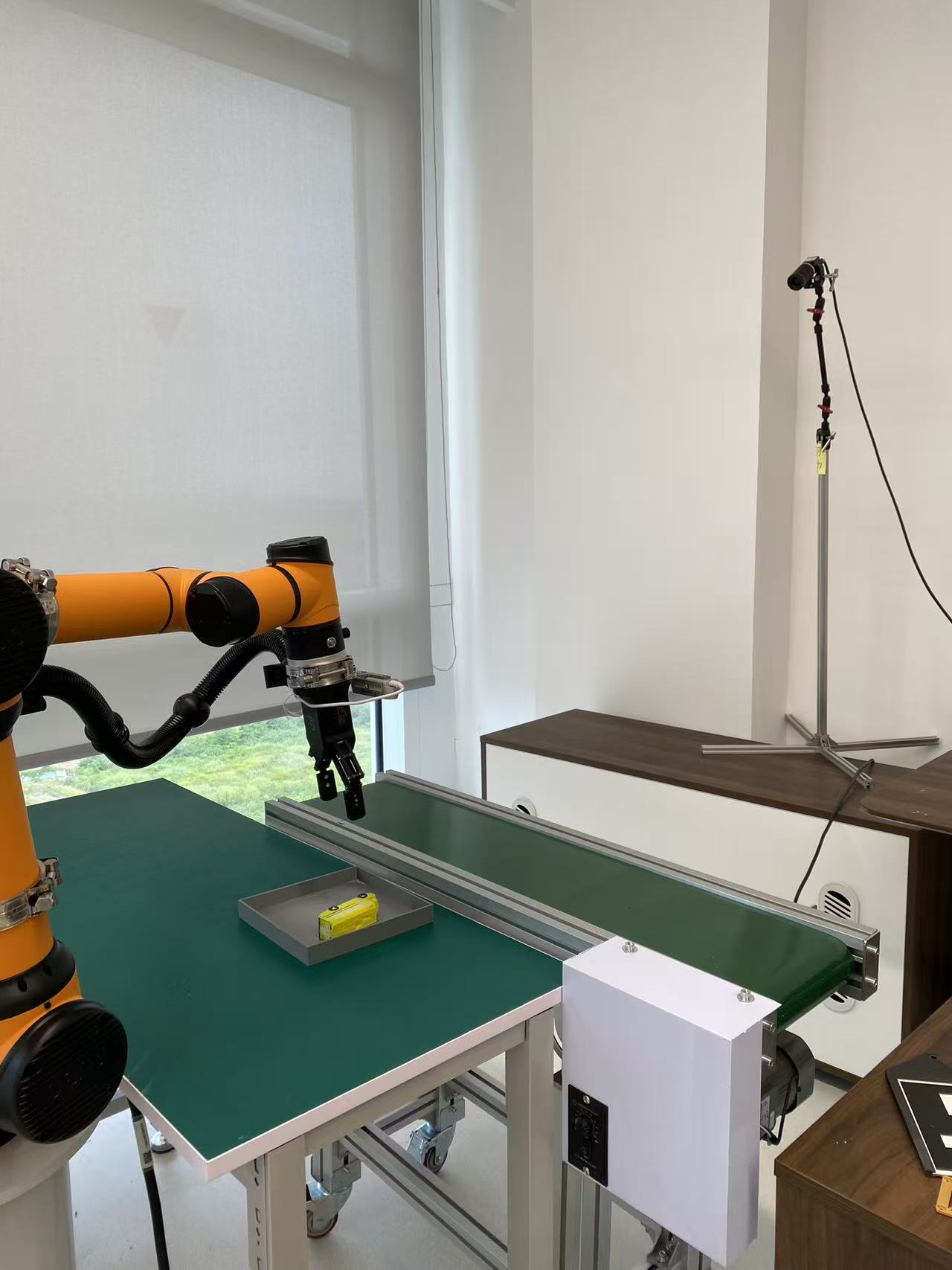}}
            \put(0.23,0.82){\colorbox{white}{\scriptsize HIKON camera}}
            \linethickness{0.8pt}
            \put(0.62,0.84){\vector(3,1){0.22}}
        \end{picture}
        \endgroup
        \par\scriptsize AUBOi5H--CTAG2F90D workcell
    \end{minipage}\hspace{0.02\linewidth}%
    \begin{minipage}[t]{0.42\linewidth}
        \vspace{0pt}
        \begin{minipage}[t]{0.48\linewidth}
            \vspace{0pt}\centering
            \parbox[c][0.98\linewidth][c]{\linewidth}{\centering
                \includegraphics[width=\linewidth,trim=80bp 400bp 40bp 400bp,clip]{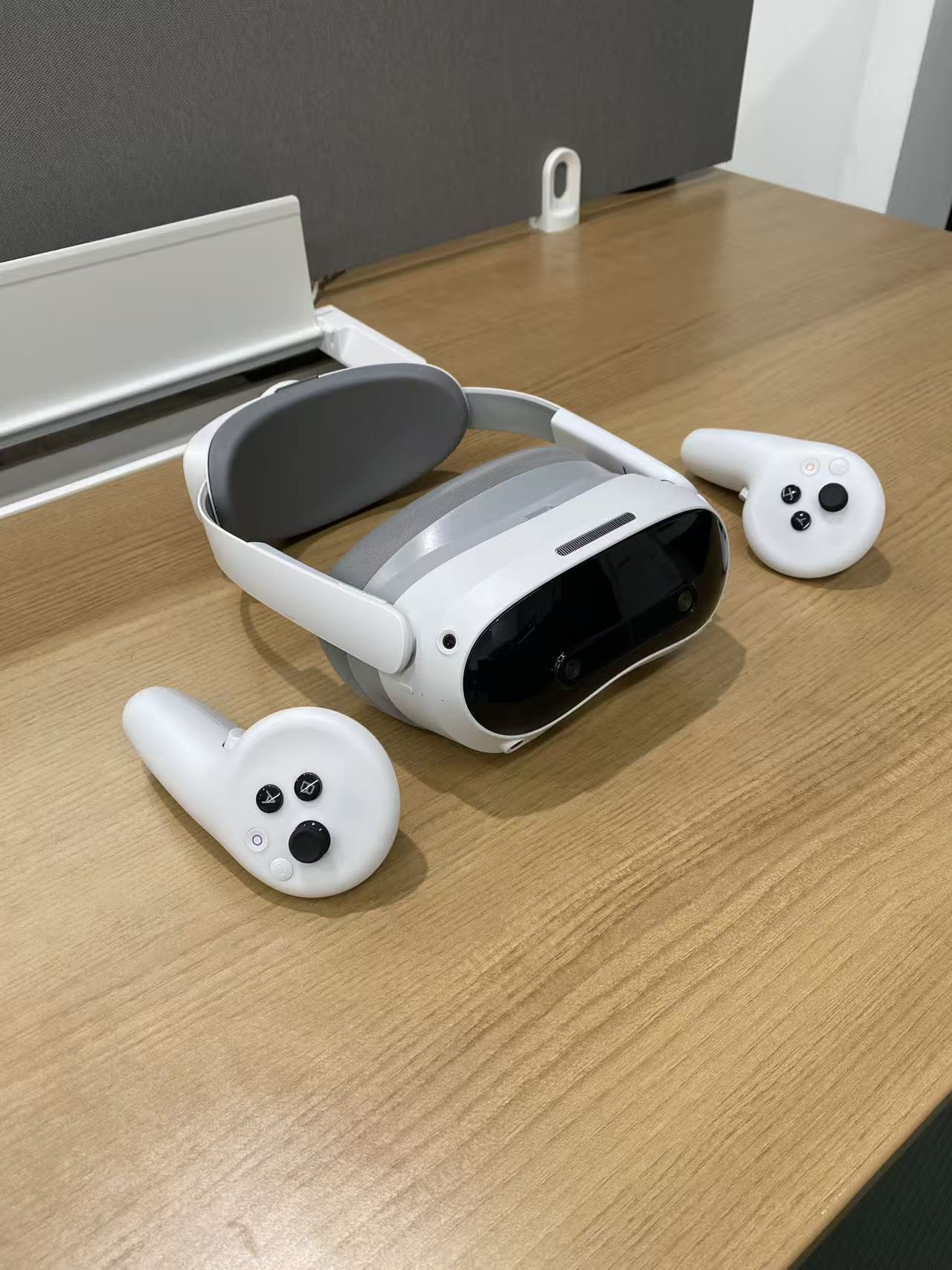}}
            \par\scriptsize Pico equipment
        \end{minipage}\hfill
        \begin{minipage}[t]{0.48\linewidth}
            \vspace{0pt}\centering
            \parbox[c][0.98\linewidth][c]{\linewidth}{\centering
                \includegraphics[width=\linewidth]{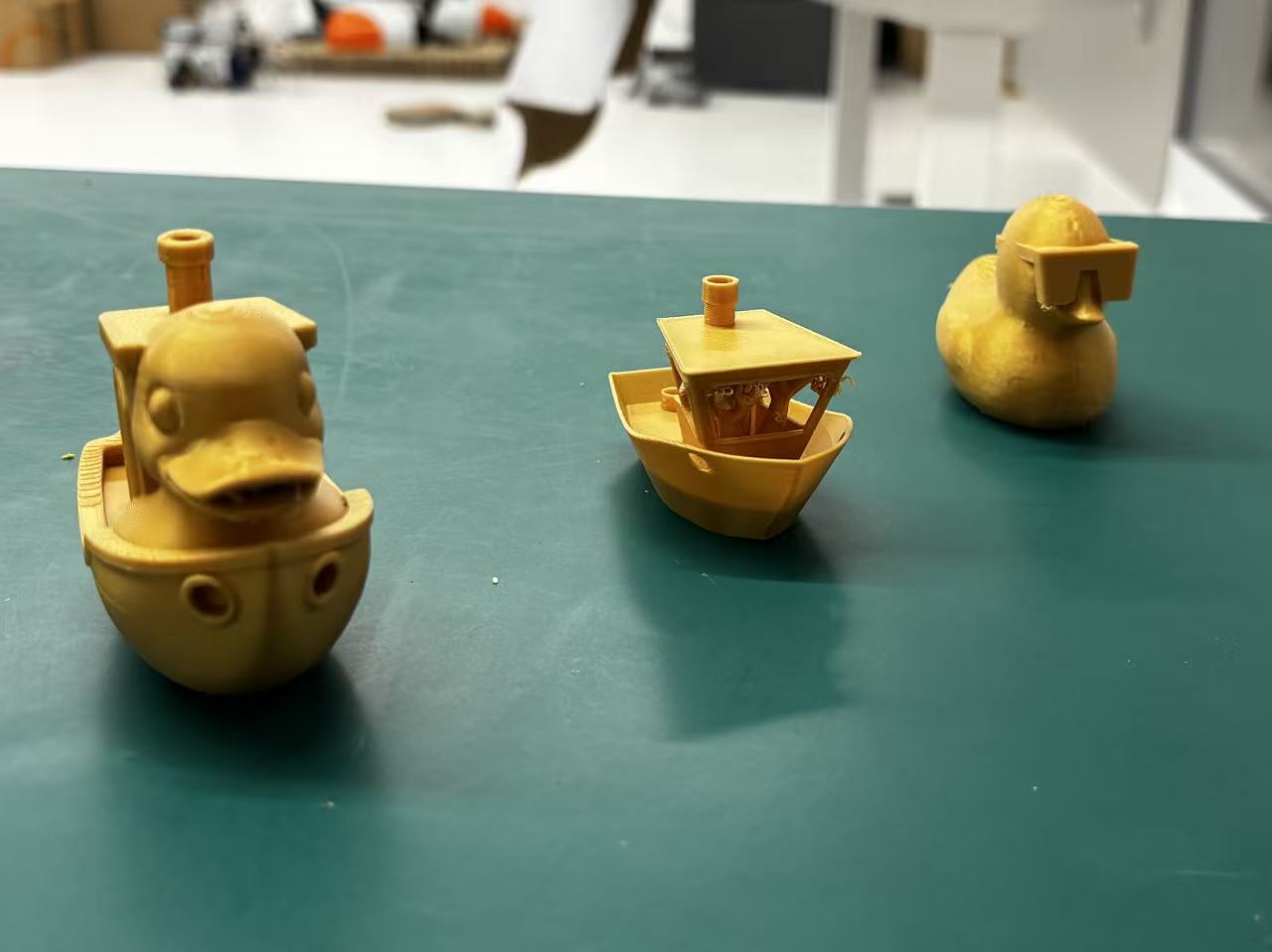}}
            \par\scriptsize OOD objects
        \end{minipage}
        \par\vspace{0.4em}
        \begin{minipage}[t]{0.48\linewidth}
            \vspace{0pt}\centering
            \includegraphics[width=\linewidth,trim=0bp 150bp 0bp 300bp,clip]{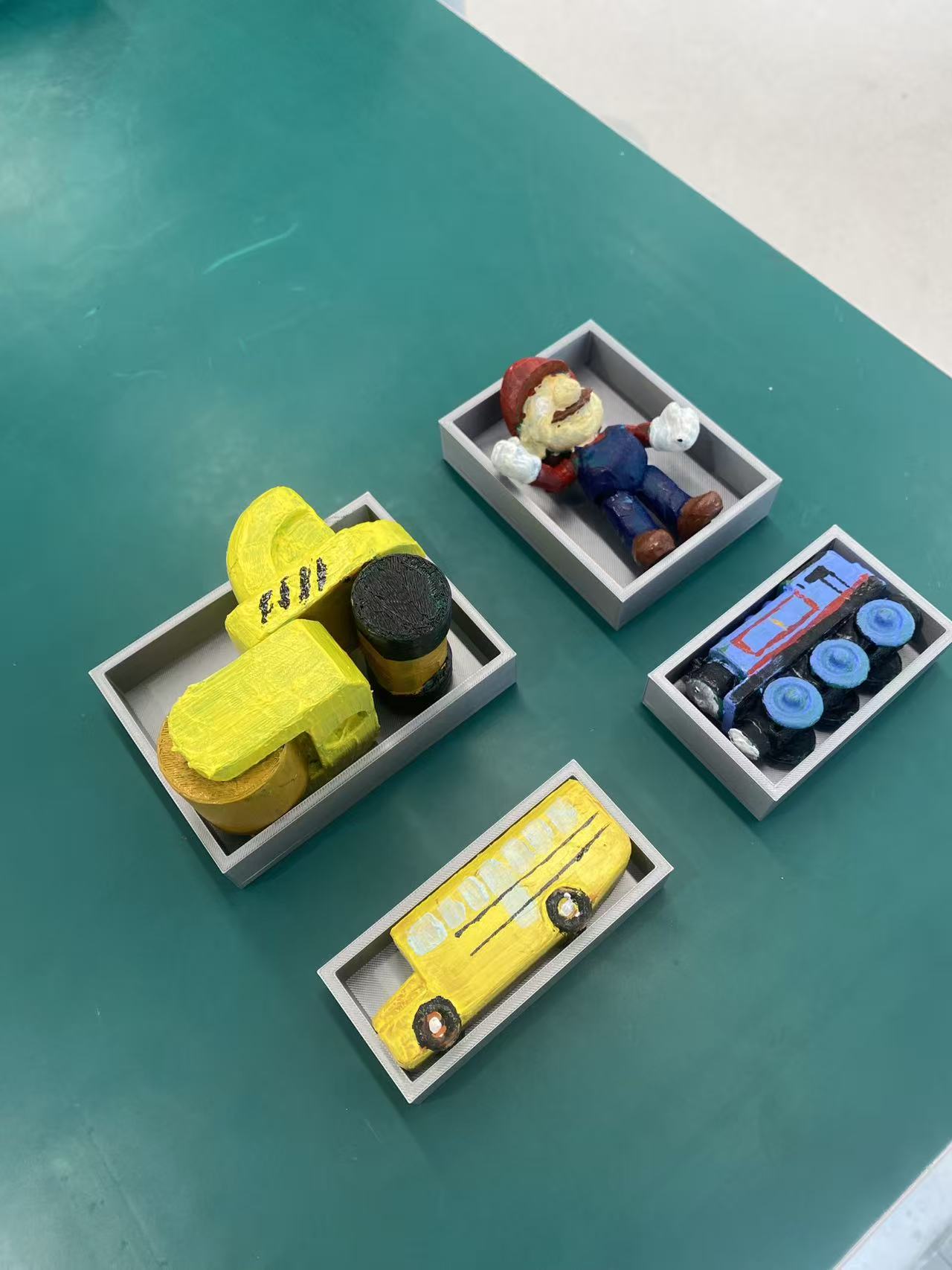}
            \par\scriptsize Level~2 grooves
        \end{minipage}\hfill
        \begin{minipage}[t]{0.48\linewidth}
            \vspace{0pt}\centering
            \includegraphics[width=\linewidth,trim=0bp 150bp 0bp 300bp,clip]{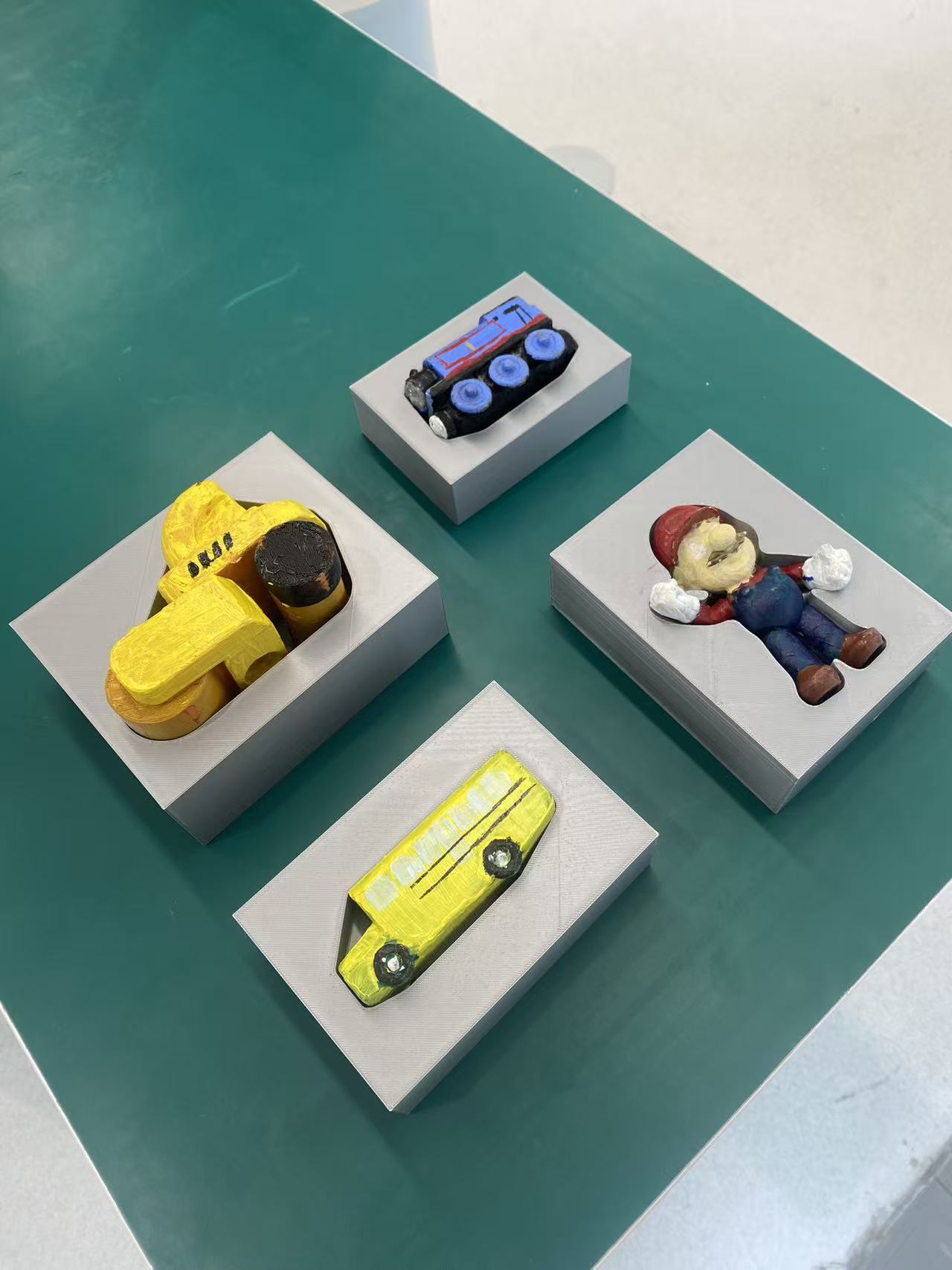}
            \par\scriptsize Level~3 inlays
        \end{minipage}
    \end{minipage}
    \caption{\textbf{Physical D2S workcell and fixtures.} Left: AUBOi5H--CTAG2F90D workcell with conveyor, target tray, and marked HIKON camera. Right, reading order: Pico teleoperation equipment, example OOD objects, Level~2 grooves, and Level~3 inlays.}
    \label{fig:rw_physical_setup}
\end{figure}
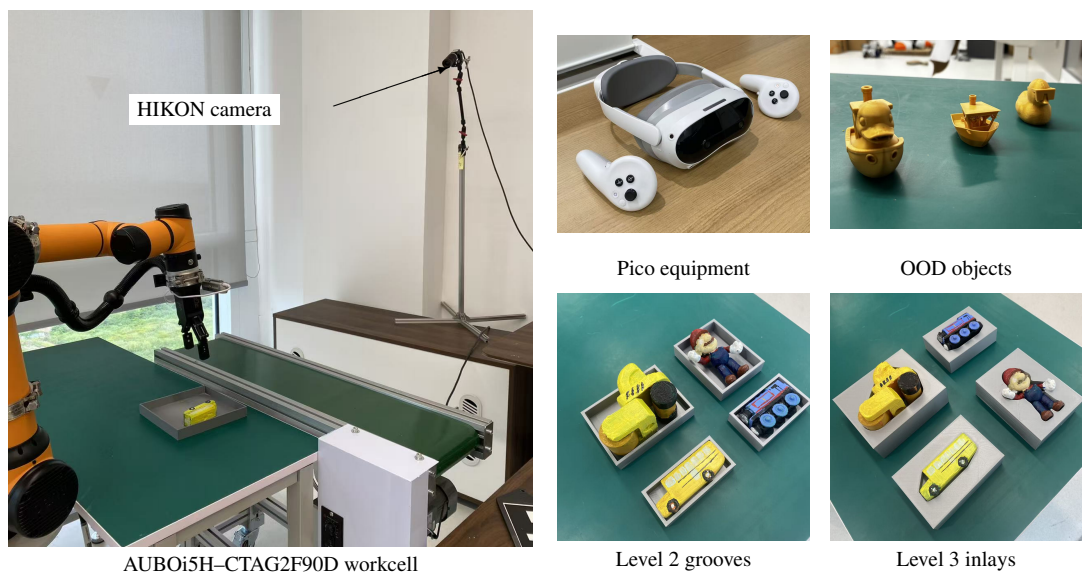

\subsection{Real-World Collection, Splits, and Quality Control}
\label{app:real_world_dataset}

The physical study tests D2S transfer in the same robot and camera workcell
across three levels. Level~1 uses tray targets; Level~2 introduces an
initial-region or target-region confounder and uses matching grooves; Level~3
combines initial- and target-region confounders with form-fitting inlays. The
levels probe pickup under scene variation and increasingly constrained target
placement within the same conveyor-to-stationary task structure. Because these
factors change together, level differences do not isolate their individual
effects.
The physical collection design uses four three-object combinations. For each
combination, four slot permutations are assigned to training and two disjoint
permutations are reserved for held-out evaluation, as summarized in
Table~\ref{tab:real_world_permutation_split}. A permutation specifies the
left-to-right object order in the target slots. The planned training design
contains

\begin{equation}
\begin{aligned}
&4\ \text{combinations}\times4\ \text{training permutations}\\
&\quad\times2\ \text{directions}\times3\ \text{disturbance levels}
=96\ \text{task configurations}.
\end{aligned}
\label{eq:real_world_training_design}
\end{equation}

Conveyor speed is balanced across these configurations rather than fully
crossed with every factor. A task configuration is distinct from a recorded
trajectory; the accepted demonstration counts are given in
Appendix~\ref{app:rw_demo_examples}.

\begin{table}[t]
    \centering
    \caption{Predeclared real-world training and held-out slot permutations.
    Letters denote School Bus (A), Mini Roller (B), Mario (C), and Thomas (D).}
    \label{tab:real_world_permutation_split}
    \small
    \begin{tabular}{lll}
        \toprule
        Combination & Training permutations & Held-out permutations \\
        \midrule
        ABC & ACB, BAC, CAB, CBA & ABC, BCA \\
        ABD & ABD, BAD, BDA, DAB & ADB, DBA \\
        ACD & ACD, ADC, CDA, DCA & CAD, DAC \\
        BCD & BDC, CBD, DBC, DCB & BCD, CDB \\
        \bottomrule
    \end{tabular}
\end{table}

The held-out permutations are reserved from the physical training split.

\subsection{Physical Demonstration Collection and Recorded Views}
\label{app:rw_demo_examples}

\begin{figure}[H]
    \centering
    \includegraphics[width=0.90\linewidth]{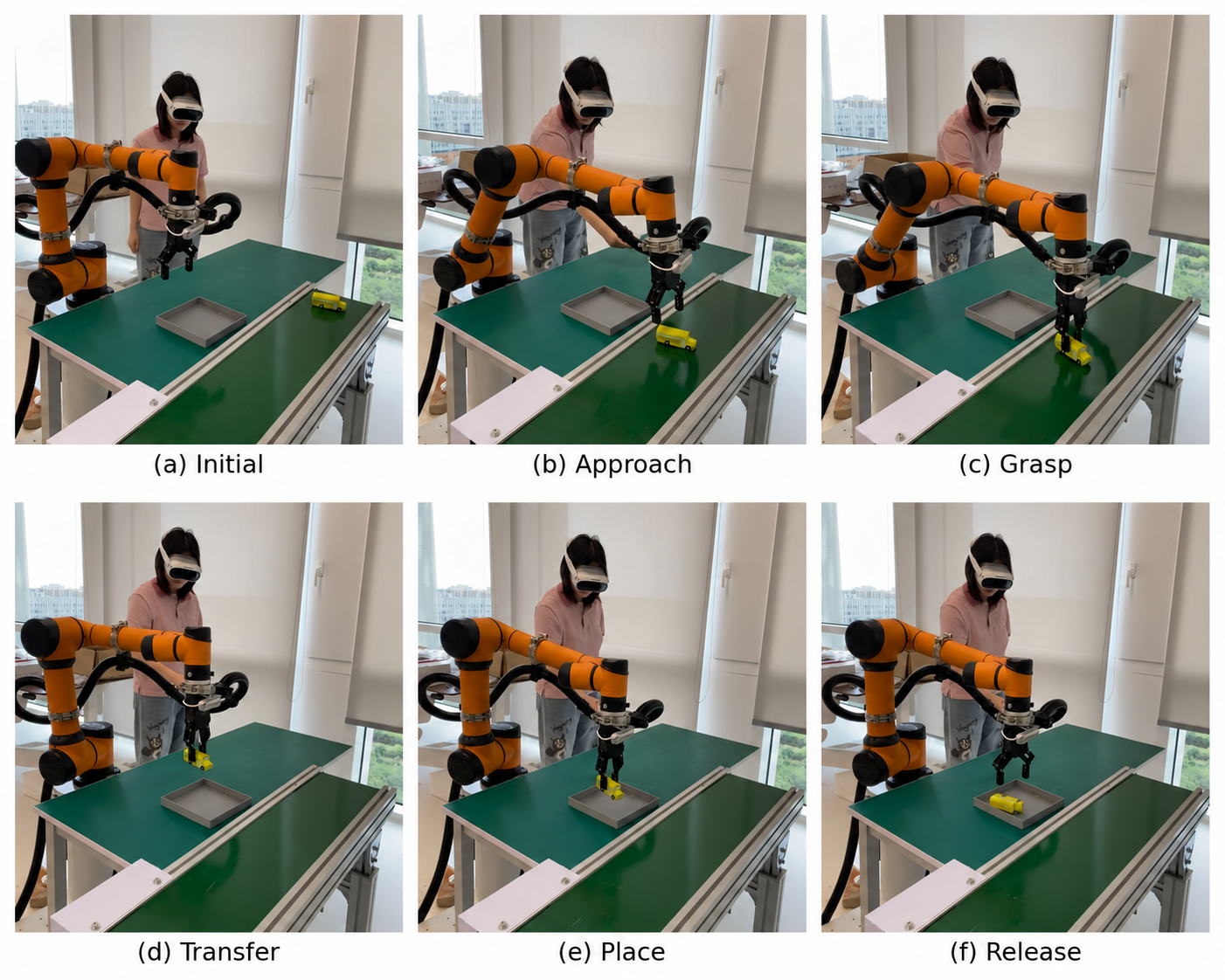}
    \caption{\textbf{Pico-assisted physical demonstration collection.} Six photographs illustrate the collection procedure: initial scene, approach, grasp, transfer, placement, and release for a Level~1 tray transfer.}
    \label{fig:rw_pico_collection}
\end{figure}

\begin{figure}[H]
    \centering
    \includegraphics[width=0.82\linewidth,trim=110bp 0 0 0,clip]{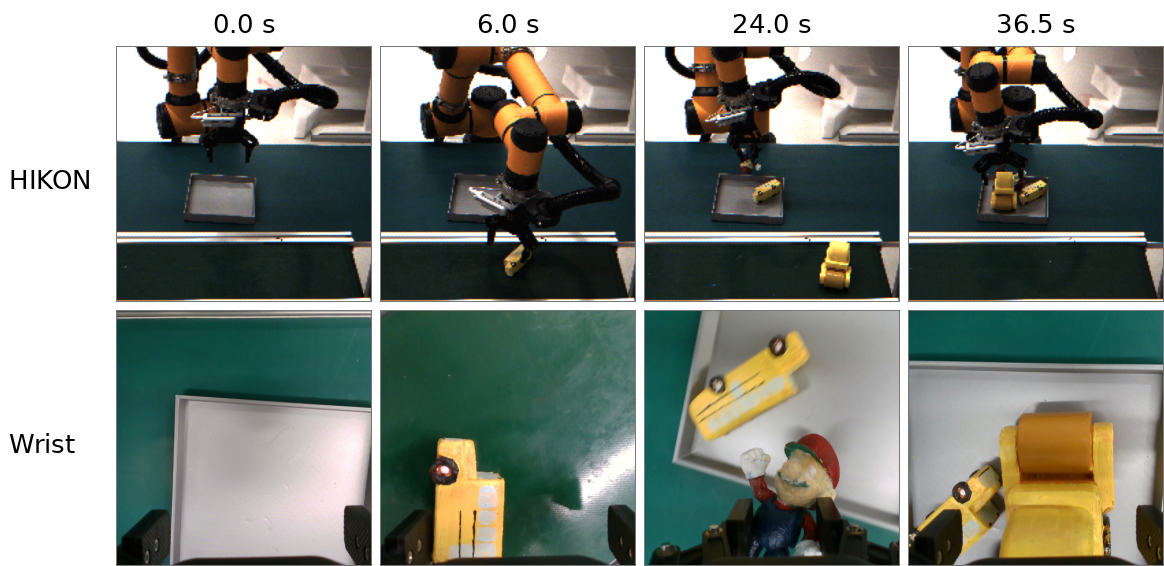}
    \caption{\textbf{Recorded physical demonstration \texttt{id\_001}.} Stored language: ``Retrieve the items from the conveyor and set them onto the plate.'' Each column pairs the table-camera view (top) with the wrist-camera view (bottom) at the same observation index of the aligned 10-Hz HDF5 episode. Times are relative to its first observation.}
    \label{fig:rw_hdf5_camera_pairs}
\end{figure}

Figure~\ref{fig:rw_pico_collection} shows the Pico-assisted collection
procedure. Figure~\ref{fig:rw_hdf5_camera_pairs} shows paired camera
observations stored in an accepted HDF5 demonstration.

\paragraph{Accepted dataset.}
The accepted physical dataset contains 95 demonstrations and 41,257 frames.
The accepted counts for Levels~1, 2, and~3 are 32, 32, and~31, respectively.
Episode \texttt{id\_092} was excluded because its nearest wrist-camera frame
failed the predeclared 25\,ms synchronization threshold. Both co-training
conditions use this same accepted release.


\providecommand{\rwTODO}[1]{\textbf{[TODO: #1]}}

\subsection{Sim--Real Co-Training Configuration}
\label{app:rw_training}

\paragraph{Training conditions.}
Both runs load the same local copy of the released $\pi_{0.5}$ base weights
(\path{pi05_base/params})~\citep{physicalintelligence2025pi05} and use the
OpenPI training implementation~\citep{openpi2025}.
S1 combines AUBO-native original-LIBERO recovered-v2 simulation demonstrations with physical demonstrations.
S2 replaces the simulation component with LIBERO-Industry demonstrations.
Both conditions use the same 95 physical demonstrations.
Table~\ref{tab:rw_training_sources} summarizes the training sources.
At physical evaluation, S1 and S2 face the same D2S workcell, object
arrangements, target fixtures, and scoring rules. The comparison therefore
tests the effect of changing the simulation training source under shared
physical supervision; it does not attribute any difference to an individual
confounder or fixture.

\begin{table}[H]
\centering
\small
\begin{tabular}{@{}p{0.34\linewidth}p{0.29\linewidth}p{0.29\linewidth}@{}}
\toprule
Training source & S1 & S2 \\
\midrule
Simulation dataset
& Original-LIBERO recovered-v2
& LIBERO-Industry \\
Simulation episodes
& 1,724
& 1,412 \\
Simulation frames
& 313,102
& 620,660 \\
Physical episodes
& 95
& 95 \\
Physical frames
& 41,257
& 41,257 \\
\bottomrule
\end{tabular}
\caption{
Training sources for the two co-training conditions.
The physical dataset is shared, while the simulation source differs.
}
\label{tab:rw_training_sources}
\end{table}

\paragraph{Sampling and optimization.}
Sampling operates on LeRobot frames using a 9:1 simulation-to-real ratio.
This ratio specifies the sampling distribution rather than the proportion of unique trajectories.
Both runs are configured for 30,000 optimizer updates with a global batch size of 252.
The shared model and optimization settings are listed in Table~\ref{tab:rw_training_config}.
This comparison matches the physical-data source, sampling weights, and configured optimization budget, but not the sizes of the unique simulation datasets.

\begin{table}[H]
\centering
\small
\begin{tabular}{@{}p{0.43\linewidth}p{0.49\linewidth}@{}}
\toprule
Setting & Shared configuration \\
\midrule
Model & $\pi_{0.5}$ \\
Base checkpoint & \texttt{pi05\_base} \\
State input & Continuous \\
Action horizon & 10 \\
Simulation-to-real sampling & 90:10, frame-based \\
Global batch size & 252 \\
Configured optimizer updates & 30,000 \\
Optimizer & AdamW \\
Gradient-norm clipping & 1.0 \\
Learning-rate schedule & Cosine schedule with warmup \\
Warmup steps & 10,000 \\
Peak learning rate & $5\times10^{-5}$ \\
Decay learning rate & $5\times10^{-5}$ \\
Scheduler decay steps & 1,000,000 \\
EMA decay & 0.999 \\
Training seed & 42 \\
Training hardware & Six NVIDIA A100 GPUs \\
FSDP devices & 2 \\
Data-loader workers & 2 \\
Checkpoint save interval & 1,000 steps \\
\bottomrule
\end{tabular}
\caption{
Shared configuration specified by the S1 and S2 launch records.
The scheduler's decay-step parameter is distinct from the number of optimizer updates.
}
\label{tab:rw_training_config}
\end{table}

\end{document}